\documentclass[10pt,twocolumn,letterpaper]{article}

\usepackage[pagenumbers]{cvpr} 

\usepackage{comment}

\definecolor{cvprblue}{rgb}{0.21,0.49,0.74}
\usepackage[pagebackref,breaklinks,colorlinks,allcolors=cvprblue]{hyperref}

\usepackage[accsupp]{axessibility}  

\usepackage{amsfonts}
\usepackage{amsmath}
\usepackage{amssymb}
\usepackage{amsthm}
\usepackage{multirow}
\usepackage{array}
\usepackage{calc}
\usepackage{tabularx}
\usepackage{booktabs}
\usepackage{afterpage}
\usepackage{makecell}
\usepackage{cuted}
\usepackage{color,soul}
\usepackage[dvipsnames,table]{xcolor}
\usepackage{makecell}
\usepackage{multirow}
\usepackage{mathrsfs}
\usepackage{overpic}
\usepackage{comment}
\usepackage{tikz}
\usepackage{pgfplots}
\pgfplotsset{compat=1.18}
\usepgfplotslibrary{colormaps}

\pgfplotsset{
    colormap={plasma}{
        rgb255=(12, 7, 134)
        rgb255=(24, 6, 139)
        rgb255=(33, 5, 143)
        rgb255=(41, 5, 147)
        rgb255=(49, 4, 150)
        rgb255=(56, 4, 153)
        rgb255=(63, 3, 156)
        rgb255=(69, 3, 158)
        rgb255=(76, 2, 161)
        rgb255=(82, 1, 163)
        rgb255=(89, 1, 164)
        rgb255=(95, 0, 166)
        rgb255=(101, 0, 167)
        rgb255=(108, 0, 168)
        rgb255=(114, 0, 168)
        rgb255=(120, 1, 168)
        rgb255=(126, 3, 167)
        rgb255=(132, 5, 166)
        rgb255=(137, 8, 165)
        rgb255=(143, 13, 163)
        rgb255=(149, 17, 161)
        rgb255=(155, 23, 158)
        rgb255=(160, 27, 155)
        rgb255=(165, 31, 151)
        rgb255=(170, 36, 148)
        rgb255=(175, 40, 144)
        rgb255=(180, 45, 141)
        rgb255=(184, 50, 137)
        rgb255=(188, 54, 133)
        rgb255=(192, 59, 129)
        rgb255=(196, 63, 126)
        rgb255=(200, 68, 122)
        rgb255=(204, 72, 118)
        rgb255=(208, 77, 115)
        rgb255=(211, 81, 111)
        rgb255=(215, 86, 108)
        rgb255=(218, 90, 104)
        rgb255=(221, 95, 101)
        rgb255=(224, 100, 97)
        rgb255=(227, 104, 94)
        rgb255=(230, 109, 90)
        rgb255=(233, 114, 87)
        rgb255=(236, 120, 83)
        rgb255=(239, 125, 79)
        rgb255=(241, 130, 76)
        rgb255=(243, 135, 72)
        rgb255=(245, 141, 69)
        rgb255=(247, 146, 65)
        rgb255=(248, 152, 62)
        rgb255=(250, 157, 58)
        rgb255=(251, 163, 55)
        rgb255=(252, 169, 52)
        rgb255=(253, 175, 49)
        rgb255=(253, 181, 45)
        rgb255=(253, 187, 43)
        rgb255=(253, 193, 40)
        rgb255=(252, 199, 38)
        rgb255=(252, 206, 37)
        rgb255=(250, 213, 36)
        rgb255=(249, 219, 36)
        rgb255=(247, 226, 37)
        rgb255=(245, 233, 38)
        rgb255=(242, 240, 38)
        rgb255=(239, 248, 33)
    }
}

\newcommand{\val}[1]{%
    \pgfmathparse{int((#1 - 1.0) / (2.0 - 1.0) * 1000)}%
    \pgfplotscolormapdefinemappedcolor{\pgfmathresult}%
    \convertcolorspec{named}{mapped color}{rgb}{\tempcolor}%
    \ifdim #1 pt < 1.6 pt \color{white} \else \color{black} \fi
    \edef\applyColor{\noexpand\cellcolor[rgb]{\tempcolor}}%
    \applyColor #1%
}

\def\paperID{42096} 
\def\confName{CVPR}
\def\confYear{2026}

\title{UDAPose: Unsupervised Domain Adaptation for \\ Low-Light Human Pose Estimation}

\author{%
   {Haopeng Chen$^1$}\quad
   {Yihao Ai$^2$}  \quad
   {Kabeen Kim$^3$}  \quad
   {Robby T. Tan$^{2,4}$}  \quad
   {Yixin Chen$^{1}$}  \quad
   {Bo Wang$^1$} \\
   $^1$University of Mississippi \;
   $^2$National University of Singapore \\
   $^3$Duksung Women's University \;
   $^4$ASUS Intelligent Cloud Services (AICS)\\
  \tt\small {hchen11@go.olemiss.edu}  \; \tt\small {yihao@u.nus.edu}  \; \tt\small {kkim15@go.olemiss.edu} \\ \tt\small {robby.tan@nus.edu.sg} \; \tt\small {yixin@olemiss.edu} \; \tt\small {hawk.rsrch@gmail.com}
}

\begin{document}
\maketitle

\begin{strip}
{
\includegraphics[width=\linewidth]{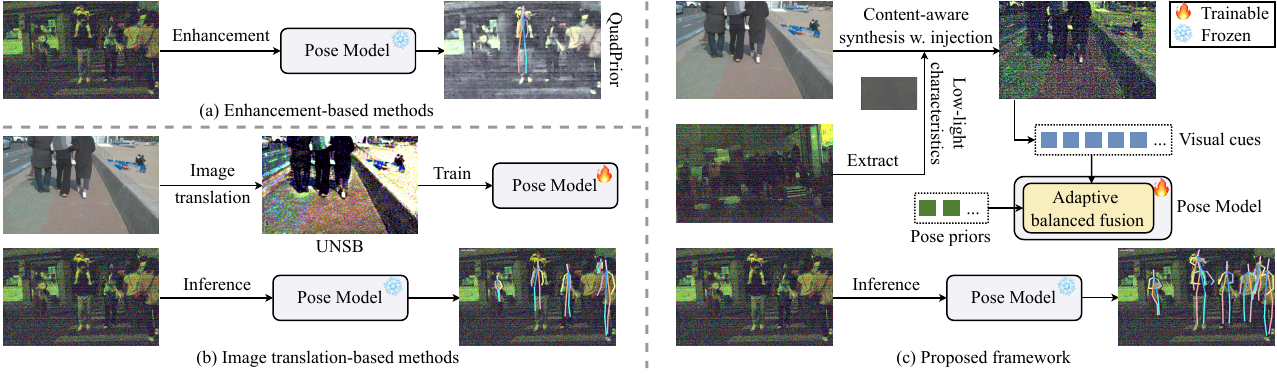}
\captionof{figure}{
Comparison of low-light human pose estimation paradigms.
(a) Image enhancement-based methods (e.g., QuadPrior~\cite{quadprior}). 
(b) Image translation-based methods (e.g., UNSB~\cite{kim2023unsb}). 
(c) Our approach synthesizes low-light image from a well-lit one with content-aware low-light characteristics injection, and balances the contribution from pose priors and image cues. 
Images are scaled for visualization only. 
}
\label{fig:representative_result}
}

\end{strip}

\begin{abstract}
Low-visibility scenarios, such as low-light conditions, pose significant challenges to human pose estimation due to the scarcity of annotated low-light datasets and the loss of visual information under poor illumination. 
Recent domain adaptation techniques attempt to utilize well-lit labels by augmenting well-lit images to mimic low-light conditions.
But handcrafted augmentations oversimplify noise patterns, while learning-based methods often fail to preserve high-frequency low-light characteristics, producing unrealistic images that lead pose models to generalize poorly to real low-light scenes.
Moreover, recent pose estimators rely on image cues through image-to-keypoint cross-attention, but these cues become unreliable under low-light conditions.
To address these issues, we propose Unsupervised Domain Adaptation for Pose Estimation (UDAPose), a novel framework that synthesizes low-light images and dynamically fuses visual cues with pose priors for improved pose estimation.
Specifically, our synthesis method incorporates a Direct-Current-based High-Pass Filter (DHF) and a Low-light Characteristics Injection Module (LCIM) to inject high-frequency details from input low-light images, overcoming rigidity or the detail loss in existing approaches.
Furthermore, we introduce a Dynamic Control of Attention (DCA) module that adaptively balances image cues with learned pose priors in the Transformer architecture.
Experiments show that UDAPose outperforms state-of-the-art methods, with notable AP gains of 10.1 (56.4\%) on the ExLPose-test hard set (LL-H) and 7.4 (31.4\%) in cross-dataset validation on EHPT-XC.
Code: \href{https://github.com/Vision-and-Multimodal-Intelligence-Lab/UDAPose}{VMIL/UDAPose}.
\end{abstract}

\section{Introduction}
\label{sec:intro}

\begin{figure*}[!t]
    \centering
    

    \begin{minipage}[c]{0.19\linewidth}
        \includegraphics[width=\linewidth]{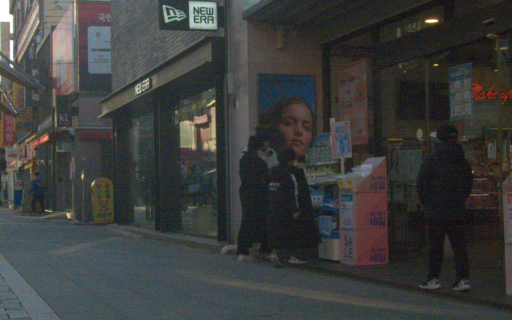}
    \end{minipage}
    \begin{minipage}[c]{0.19\linewidth}
        \includegraphics[width=\linewidth]{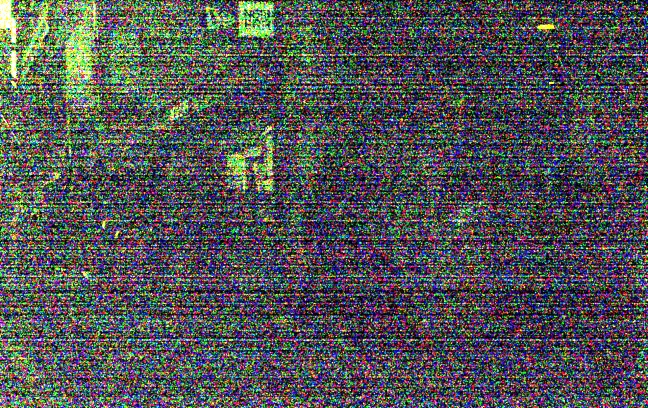}
    \end{minipage}
    \begin{minipage}[c]{0.19\linewidth}
        \includegraphics[width=\linewidth]{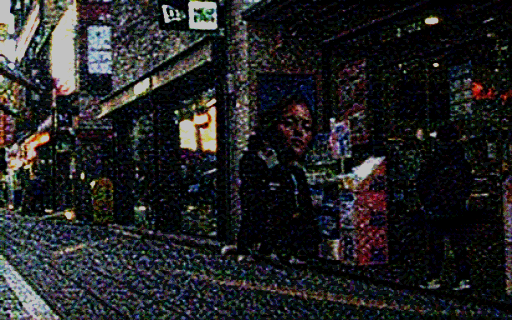}
    \end{minipage}
    \begin{minipage}[c]{0.19\linewidth}
        \includegraphics[width=\linewidth]{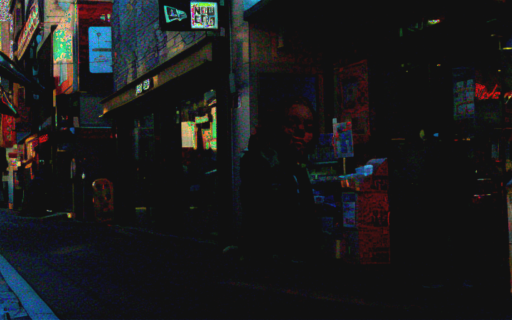}
    \end{minipage}
    \begin{minipage}[c]{0.19\linewidth}
        \includegraphics[width=\linewidth]{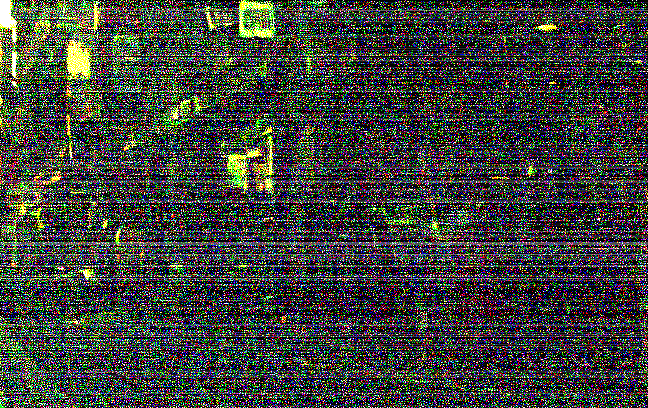}
    \end{minipage}
    %
    \begin{minipage}[c]{0.19\linewidth}
        \centering{\scriptsize Well-lit}
    \end{minipage}
    \begin{minipage}[c]{0.19\linewidth}
        \centering{\scriptsize Paired Low-light}
    \end{minipage}
    \begin{minipage}[c]{0.19\linewidth}
        \centering{\scriptsize CycleGAN}
    \end{minipage}
    \begin{minipage}[c]{0.19\linewidth}
        \centering{\scriptsize StyleID}
    \end{minipage}
        \begin{minipage}[c]{0.19\linewidth}
        \centering{\scriptsize Ours}
    \end{minipage}
    \caption{
    Limitations of learning-based low-light augmentation.
    %
    The first two columns show well-lit and paired low-light images from ExLPose~\cite{ExLPose_2023_CVPR}.
    The third and fourth columns present results from CycleGAN~\cite{CycleGAN2017} and StyleID~\cite{chung2024style}.
    The last column shows our result. 
    Low-light images are scaled to an average channel intensity of 0.4 for visualization only.
    }
    \label{fig:intro-baseline-gen-comp}
\end{figure*}
Human pose estimation is a foundational task in computer
vision, essential for many downstream applications~\cite{application-healthcare,application-traffic,application-sports2,application-ar,application-security,survey-application-2021}.
Existing methods~\cite{rw-poseur,purkrabek2025probpose,rw-swahr,LOGOCAP,grouppose,tan2024diffusionregpose}
and benchmark datasets~\cite{lin2014microsoft,li2019crowdpose,dataset-ochuman}
primarily focus on well-illuminated scenarios.
However, real-world scenarios often involve low-light conditions, which significantly degrade pose estimation performance~\cite{ExLPose_2023_CVPR,ehpt-xc} and result in safety risks~\cite{tesla-NHTSA}. 
A key challenge is the scarcity of real-world low-visibility datasets, as annotating such images is inherently difficult.
\citet{ExLPose_2023_CVPR} introduced a specialized camera system to capture paired well-lit and low-light images, transferring annotations from well-lit to low-light counterparts.
However, these artificially darkened images cannot fully replicate real
low-light conditions, limiting the generalization of models trained on them~\cite{li2021learning}.
Although event cameras~\cite{ehpt-xc} offer an alternative, they require specialized hardware and complex cross-modality alignment, which limits scalable deployment.

While paired data collection is impractical, an alternative approach is to apply low-light image enhancement methods~\cite{yang2023implicit,quadprior,cai2023retinexformer,feijoo2025darkir,jiang2024lightendiffusion}.
However, this recovery process is inherently ill-posed, as reconstructing missing visual details from severely degraded images is challenging and often leads to artifacts that negatively impact human pose estimation.\textbf{}
Instead of recovering lost details, domain-adaptive methods~\cite{ella-eccv,kim2022unified} take a different approach by synthesizing low-light images from well-lit ones to mimic low-visibility conditions.
This allows models to leverage existing well-lit annotations during training.
These methods typically rely on handcrafted augmentations~\cite{ella-eccv} or learning-based image-to-image translation~\cite{CycleGAN2017,cai2024enco,kim2023unsb} to bridge the domain gap.
However, their effectiveness depends on how well they replicate real low-light characteristics, which remains a significant and underexplored challenge.

Handcrafted augmentations often fail to replicate the complex characteristics of real low-light images.
For instance, ELLA~\cite{ella-eccv} applies Gaussian white noise to simulate low-light conditions. 
However, real low-light noise, such as photon noise, thermal noise, and quantization noise, is far more complex~\cite{wei2021physics}. 
%
Moreover, handcrafted augmentations exhibit limited flexibility and generalization, as they are tailored to specific low-light scenarios. 
Consequently, their deployment in novel environments, such as those involving different camera hardware or new datasets (e.g., EHPT-XC~\citep{ehpt-xc}), often results in suboptimal performance and requires extensive manual tuning.
Learning-based augmentations utilize unpaired image-to-image translation~\cite{CycleGAN2017, kim2023unsb, chung2024style} to adapt well-lit images to low-light conditions.
However, as shown in~\cref{fig:intro-baseline-gen-comp}, these methods fail to replicate realistic low-light characteristics. 
CycleGAN~\cite{CycleGAN2017} tends to overly darken images while introducing lighting artifacts, while StyleID~\cite{chung2024style} fails to generate realistic low-light noise.

Beyond the limitations of low-light data synthesis, the robustness of modern pose estimators themselves becomes a critical factor, particularly the recent one-stage architectures~\cite{edpose,grouppose,tan2024diffusionregpose}.
When visual cues are subtle or entirely absent, a robust model should leverage learned pose priors to infer keypoints hidden in darkness.
However, recent one-stage pose estimators~\cite{edpose,grouppose,tan2024diffusionregpose}, often built upon DETR-like architectures~\cite{DETR,deformable-detr}, utilize cross-attention to query image features and fuse them with pose priors via a direct residual connection. 
This rigid summation biases the fused representation toward image cues, even when the visual information is unreliable, particularly under low-light conditions.
Empirically, we observe that the model consistently emphasizes cross-attention visual features over pose-prior features under both well-lit and low-light conditions (see~\cref{fig:l2norm} for L2-norm comparisons).
Consequently, under poor illumination, noisy and unreliable visual features still contribute significantly compared to the pose priors, leading to unreliable human pose predictions, particularly for the low-visibility keypoints.

To overcome the above limitations, we propose UDAPose, a novel framework for unsupervised domain adaptation in human pose estimation.
Our framework employs Stable Diffusion (SD)~\cite{latentdiffusion} as a generative backbone to synthesize low-light images from well-lit ones.
By using unlabeled low-light images as references, UDAPose synthesizes augmentations that better reflect real low-light characteristics, outperforming existing approaches.
Our approach is highly practical, as collecting unlabeled low-light images is far easier than obtaining corresponding pose annotations.
To capture low-light characteristics from reference images, we introduce (1) Direct-Current-based High-Pass Filter (DHF), which extracts high-frequency low-light characteristics and (2) Low-Light Characteristics Injection Module (LCIM), which ensures that synthesized images retain complex low-light features of the reference images.
Unlike existing learning-based augmentations, UDAPose preserves essential low-light characteristics for human pose estimation, achieving significant performance gains.
A representative example is shown in the last column of~\cref{fig:intro-baseline-gen-comp}.

To address the vulnerability of pose estimation models under low-light conditions, we introduce the Dynamic Control of Attention (DCA) module. 
DCA adaptively controls the fusion weight between visual cues and pose priors, allowing the model to dynamically adjust the contributions from image features and learned human pose priors when visual information is degraded.
In summary, our contributions are as follows. 
\begin{itemize}
    \item 
    We propose UDAPose, an unsupervised domain adaptation framework that augments well-lit images to mimic low-light conditions and adaptively fuses visual cues with learned pose priors, achieving improved low-light human pose estimation without requiring low-light annotations.
    \item 
    We introduce the Direct-Current-based High-Pass-Filter (DHF) and Low-Light Characteristics Injection Module (LCIM), which preserve and inject high-frequency details to the synthesized low-light images.
    \item
    We propose Dynamic Control of Attention (DCA), which dynamically balances image cues with learned pose priors in the Transformer architecture, reducing the influence of noisy and unreliable visual cues in low-light scenarios. 
\end{itemize}
Experimental results show UDAPose surpasses existing methods, achieving a 10.1 AP (56.4\%) improvement on the low-light hard set (LL-H) of ExLPose-test and a 7.4 AP (31.4\%) improvement in cross-dataset evaluation on EHPT-XC, highlighting its robustness under low-light conditions.

\section{Related Work}
\label{sec:related}
\noindent\textbf{Human Pose Estimation}
Modern human pose estimation is primarily categorized into two mainstream paradigms: top-down and bottom-up. Top-down approaches~\cite{rw-poseur,wang2024locllm,purkrabek2025probpose} first detect individuals and subsequently estimate the pose for each one. In contrast, bottom-up approaches~\cite{DEKR2021,rw-swahr,LOGOCAP} first detect all body keypoints in the image and then assemble them into distinct person instances.
Recent advances, including one-stage unified detection-estimation frameworks~\cite{tian2019directpose,mao2021fcpose,shi2021inspose,xiao2022querypose,edpose,grouppose,tan2024diffusionregpose} and Vision Transformer-based models~\cite{rw-petr,xu2022vitpose,wang2024locllm,khirodkar2024sapiens}, have achieved remarkable accuracy. 
However, these models are trained and evaluated primarily on benchmarks with ideal lighting conditions~\cite{lin2014microsoft,li2019crowdpose,dataset-lsp,dataset-ochuman}. In particular, recent one-stage methods~\cite{edpose,grouppose,tan2024diffusionregpose} that rely on image cues become unreliable under low-light conditions.
Consequently, their performance degrades significantly in low-light scenarios~\cite{ExLPose_2023_CVPR,ehpt-xc}. 
While some works have explored domain adaptation~\cite{kim2022unified, peng2023source, raychaudhuri2023prior, cao2019cross, li2021synthetic, mu2020learning, han2022learning, jiang2021regressive, jin2022multibranch} or simple rule-based augmentations~\cite{ella-eccv}, they either lack specific designs for low-light characteristics or fail to synthesize realistic degradation. 
Methods requiring paired low-light and well-lit images~\cite{ExLPose_2023_CVPR}, or additional event camera data~\cite{ehpt-xc} are impractical due to the difficulty of collecting such data at scale and their limited scalability in deployment.

\noindent\textbf{Low-light Image Enhancement}
An intuitive approach for low-light pose estimation is to first apply a low-light image enhancement (LLIE) method as a pre-processing step. 
While deep learning techniques, including CNNs~\cite{moran2020deeplpf, sharma2021nighttime, wang2019underexposed}, GANs~\cite{huang2017arbitrary, jin2022unsupervised, jiang2021enlightengan, yang2023implicit}, Transformers~\cite{cai2023retinexformer,feijoo2025darkir}, and diffusion models~\cite{LLFlow, jiang2024lightendiffusion,quadprior}, have surpassed traditional methods like histogram equalization~\cite{celik2011contextual, cheng2004simple} and Retinex-based approaches~\cite{li2018structure, wang2013naturalness}, they still face significant challenges. 
These methods can introduce artifacts or fail to restore sufficient detail in extremely dark images, which in turn limits the performance of any subsequent pose estimation model.

\noindent\textbf{Unpaired Image-to-Image Translation}
Unpaired image-to-image (I2I) translation offers a promising direction for generating synthetic training data. 
Seminal works like CycleGAN~\cite{CycleGAN2017} enabled translation without paired data, a concept extended by style transfer methods like AdaIN~\cite{huang2017arbitrary}. 
More recently, diffusion-based models~\cite{kim2023unsb,chung2024style,xia2024diffusion,xia2024diffi2i} have demonstrated powerful capabilities in domain mapping. 
However, the primary objective of these generic I2I methods is to alter global appearance and texture. They are not specifically designed to synthesize the complex, non-uniform noise and structural degradation characteristic of low-light conditions, which is crucial for training a robust pose estimator. 
This distinction motivates our work on a specialized low-light augmentation strategy tailored for high-level vision tasks.
\begin{figure*}[!ht]
    \centering
    \includegraphics[trim={0 0.35cm 0.5cm 0},clip,width=\textwidth]{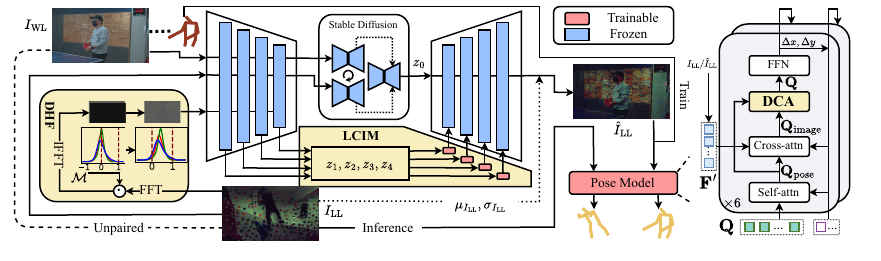}
\caption{
    Overview of the UDAPose framework. 
    During augmentation, the LCIM uses extracted low-light features from unpaired low-light images ($I_\text{LL}$) to synthesize low-light counterparts ($\hat{I}_\text{LL}$) of well-lit images ($I_\text{WL}$).
    These synthetic images retain the original pose annotations from $I_\text{WL}$ while accurately reflecting low-light characteristics of $I_\text{LL}$.
    The pose model is trained using these augmented images and their inherited annotations.
    Our DCA adaptively balances image cues and pose priors under low-light conditions.
    During inference, the trained model is directly applied to real low-light images. Note that $I_\text{LL}$, $I'_\text{LL}$, and $\hat{I}_\text{LL}$ are scaled for visualization only. 
}
\label{fig:overall_arch}
\end{figure*}

\section{Method}
\label{sec:method}
\noindent\textbf{Overview} 
UDAPose is a novel unsupervised domain adaptation framework that synthesizes low-light images from well-lit ones and adaptively fuses visual cues with learned pose priors for robust human pose estimation. 
As illustrated in~\cref{fig:overall_arch}, our approach uses a pre-trained Stable Diffusion (SD) model to transfer scene structure from an annotated well-lit image while injecting low-light characteristics from an unlabeled reference image, enabling supervised pose training in the low-light conditions without requiring low-light annotations. 
We obtain the style-infused latent code $z_0$, which embeds the structure of the well-lit image $I_\text{WL}$ and the low-frequency style of the reference $I_\text{LL}$ following~\citet{chung2024style}.
During the decoding process, $z_0$ is decoded, where low-light noise patterns are injected through two key modules, DHF and LCIM.
The diffusion backbone is responsible for preserving the scene structure and low-frequency appearance from the well-lit image, while DHF and LCIM focus on injecting high-frequency low-light characteristics.
The obtained low-light images are then used as training data for our Transformer-based pose estimator, where our DCA module adaptively balances image cues and pose priors to mitigate unreliable visual information under low-light conditions.

\noindent\textbf{Direct-Current-based High-Pass Filter (DHF)}
Our framework extracts high-frequency details by applying a high-pass filter in the frequency domain, yielding an image $I_{\text{HP}}$. It can be formulated as:
\begin{equation}
    I_{\text{HP}} = \text{iFFT}(\text{FFT}(I_\text{LL}) \odot \mathcal{M} )
\end{equation}
where $\mathcal{M}$ is a high-pass filter, and $\text{FFT}$ and $\text{iFFT}$ denote the Fast Fourier Transform and its inverse. 
By design, $I_{\text{HP}}$ has a mean near zero, with its pixel values representing positive (brighter) and negative (darker) deviations from the local average.
A critical challenge arises when preparing $I_{\text{HP}}$ for the SD encoder, which expects inputs normalized to the standard RGB $[0, 1]$ range.
Direct clipping of negative values leads to irreversible information loss, especially for darker details, a critical issue in low-light scenarios. 

To address this issue, we introduce a simple yet effective module, DHF.
The core idea is to re-center high-frequency details by aligning their distribution with the mean brightness of the original reference image, $I_\text{LL}$.
This process preserves the full dynamic range of the extracted details by shifting them into a perceptually meaningful range prior to normalization. 
Specifically, we compute the corrected high-frequency image $I_{\text{DHF}}$ as:
\begin{equation}
    I_{\text{DHF}} = I_{\text{HP}} + (\text{mean}(I_\text{LL}) - \text{mean}(I_{\text{HP}})),
\end{equation}
where $\text{mean}(\cdot)$ calculates the global channel-wise mean of an image.
This operation ensures that $\text{mean}(I_{\text{DHF}}) = \text{mean}(I_\text{LL})$.
By adjusting the negative-valued details, this process reduces information loss during the final clipping to the $[0, 1]$ range. 
Consequently, DHF helps preserve both bright and dark high-frequency information, enabling a richer feature representation for low-light domain adaptation.

\noindent\textbf{Low-light Characteristic Injection Module (LCIM)}
The synthesis process is finally completed in the decoding stage, which generates the low-light image $\hat{I}_{\rm LL}$.
A variational autoencoder (VAE) decoder $\mathcal{D}$ takes the style-infused latent code $z_0$, embedding the structure of the well-lit image $I_{\rm WL}$ and the low-frequency style of the reference low-light image $I_{\rm LL}$, and progressively upsamples it.
To inject high-frequency details from the reference image into the synthesized low-light output, we introduce LCIM.

After obtaining a set of high-frequency intermediate features, ${z_1,...,z_4}$ from different scales of the high-frequency image $I_\text{DHF}$ produced by DHF, LCIM processes each $z_i$ with a lightweight convolutional layer:
\begin{equation}
    \{f_1, f_2, f_3, f_4\} = \text{LCIM}(\{z_1, z_2, z_3, z_4\}),
\end{equation}

The decoder $\mathcal{D}$ is composed of 4 convolution blocks $d_1,...,d_4$ and a convolution layer $d_\text{final}$. Each processed high-frequency feature ${f_1,...,f_4}$ is injected at the end of each convolution block by adding it to the main stream:
\begin{equation}
    \hat{I}'_\text{LL} \leftarrow d_\text{final}(d_4( d_3( d_2( d_1( z_0 ) + f_1) + f_2) + f_3) + f_4)
\end{equation}

This multi-scale injection strategy guides the synthesis process to render fine-grained low-light noise at appropriate spatial resolutions.
Finally, to match the global stylistic appearance of the reference, we align the channel-wise mean and standard deviation of the synthesized image $\hat{I}'_{LL}$ with those of $I_{\rm LL}$ to produce our final output, $\hat{I}_{\rm LL}$.

To optimize LCIM, we freeze the encoder and decoder of the VAE, and train the module to reconstruct low-light images using a composite loss that incorporates both spatial and frequency domains:
\begin{equation}
  \mathcal{L_\mathcal{D}} = \mathcal{L}_{\text{MSE}}(I, \hat{I}) + \lambda \mathcal{L}_{\text{freq}}(I, \hat{I}),
  \label{equ:llcd_loss_revised}
\end{equation}
where $I$ is the low-light input and $\hat{I}$ is its reconstruction. The first term, $\mathcal{L}_{\text{MSE}}$, is the mean squared error, which enforces content fidelity by minimizing pixel-wise differences. The second term, $\mathcal{L}_{\text{freq}}$, is a frequency-domain loss designed to preserve fine-grained details specific to low-light conditions. 
The hyperparameter $\lambda$ balances the two loss terms.
In particular, the second loss is defined as a weighted MSE on the Fourier magnitude spectra:
\begin{equation}
  \label{eq:freq_loss_revised}
  \mathcal{L}_{\text{freq}} = \frac{1}{MN}\sum_{u=0}^{M-1} \sum_{v=0}^{N-1} \mathcal{W}(u,v) |\mathcal{F}_{I}(u,v) - \mathcal{F}_{\hat{I}}(u,v)|^{2},
\end{equation}
where $\mathcal{F}_{I}$ and $\mathcal{F}_{\hat{I}}$ are the Fourier magnitude spectra of the input and reconstructed images, respectively, and $(M,N)$ is the image resolution. The weighting function $\mathcal{W}(u,v)$ is defined as:
\begin{equation}
  \label{eq:weighting_revised}
  \mathcal{W}(u,v) = \sin{\left(\frac{\pi|2u-M|}{2M}\right)} + \sin{\left(\frac{\pi|2v-N|}{2N}\right)}.
\end{equation}
This sinusoidal weighting scheme prioritizes mid-to-high frequency components, aiming to enhance perceptual quality and retain low-light characteristics without introducing over-sharpening artifacts. 
Although LCIM is trained with a reconstruction objective, it operates on high-frequency components extracted from low-light images, which mainly capture noise patterns. Thus, LCIM captures transferable low-light characteristics that can be applied to synthesize low-light effects for different input images.

\noindent\textbf{Dynamic Control of Attention (DCA)}
\begin{figure}[!t]
  \centering

    \begin{subfigure}[t]{0.58\columnwidth}
        \centering
        \includegraphics[trim={0 0 0 0.5cm},clip,width=\textwidth]{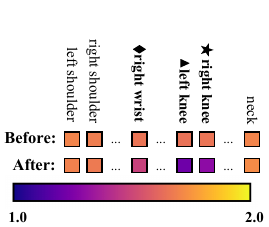}
    \caption{$\|\mathbf{Q}_\text{image}\|_2 / \| \mathbf{Q}_\text{pose}\|_2$}
    \label{fig:l2norm-vis}
    \end{subfigure}
    \hfill
    \begin{subfigure}[t]{0.18\columnwidth}
        \centering
        \includegraphics[trim={9cm 3.5cm 14.5cm 1cm},clip,width=\textwidth]{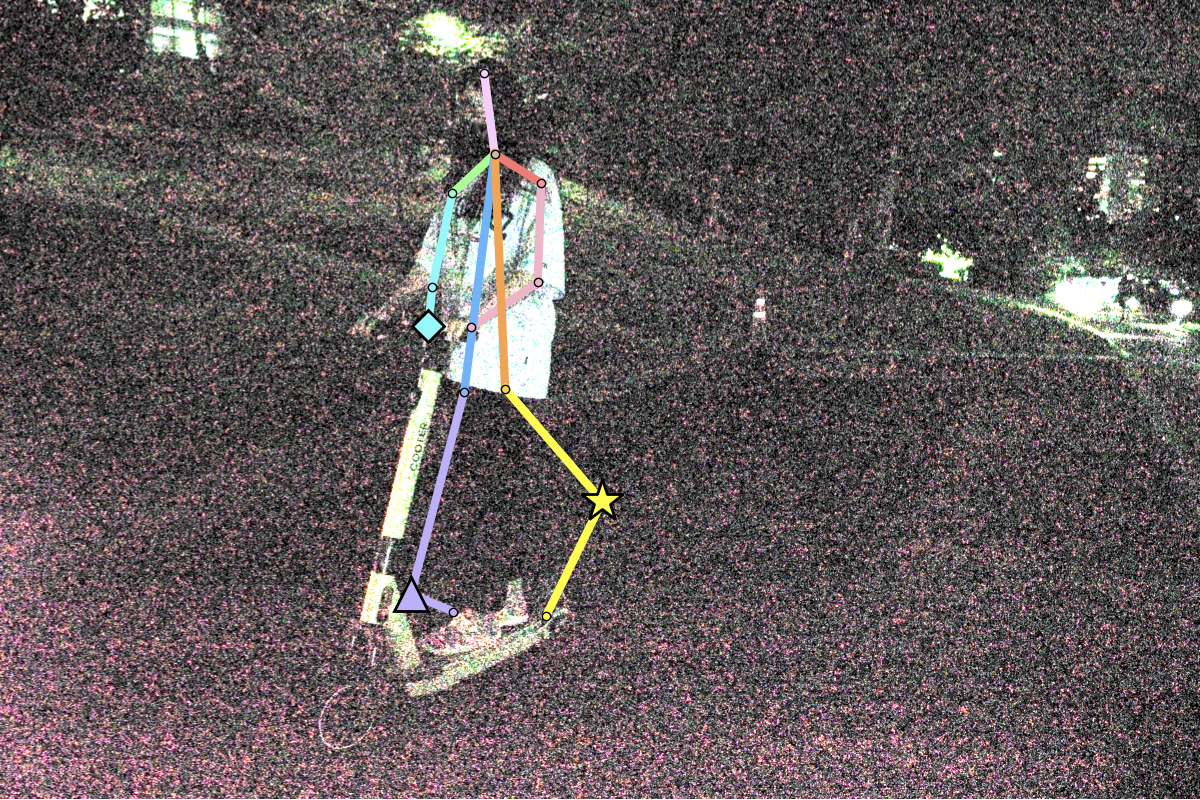}
    \caption{Before}
    \label{fig:l2norm-before}
    \end{subfigure}
    \hfill
    \begin{subfigure}[t]{0.18\columnwidth}
        \centering
        \includegraphics[trim={9cm 3.5cm 14.5cm 1cm},clip,width=\textwidth]{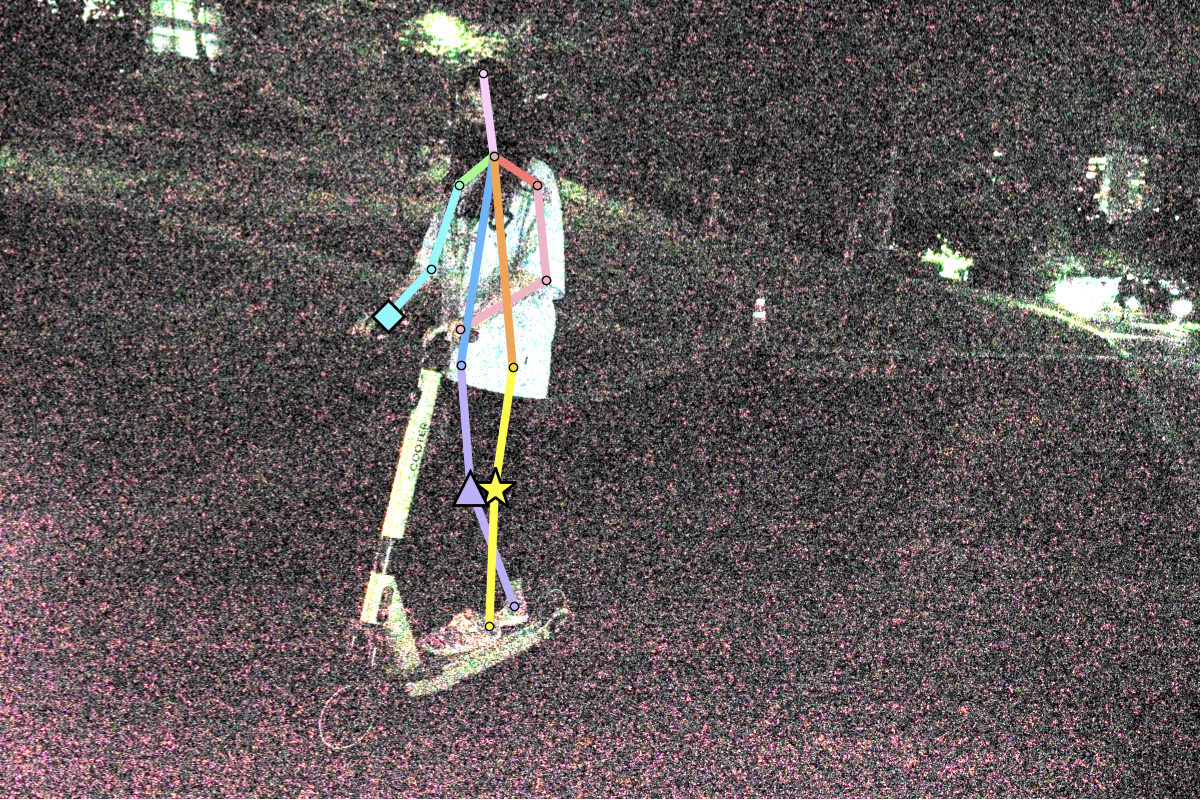}
    \caption{After}
    \label{fig:l2norm-after}
    \end{subfigure}
  
\caption{
    Ratio of Frobenius Norm of $\mathbf{Q}_\text{image}$ over $\mathbf{Q}_\text{pose}$ (i.e. $\|\mathbf{Q}_\text{image}\|_2 / \| \mathbf{Q}_\text{pose}\|_2$) on different keypoints and pose estimation results before/after applying DCA module. Note that images are scaled for visualization only.
}
\label{fig:l2norm}
\end{figure}
After synthesizing the low-light image $\hat{I}_\text{LL}$, we use it for our pose model training.
$\hat{I}_\text{LL}$ is first processed with a feature extractor and then transformed into tokens $\mathbf{F'}$ by a Transformer encoder.
We denote the visual tokens carrying image cues after deformable cross-attention as $\mathbf{Q}_\text{image}$ (corresponding to the output of deformable token-to-human/keypoint attention in \cite{edpose}).
Our pose model also initializes keypoint latents and performs self-attention among them, forming a pose-prior representation denoted as $\mathbf{Q}_\text{pose}$ (corresponding to the output of human-to-keypoint interactive attention in \cite{edpose}).
Within each deformable decoder layer, we need to fuse the pose prior latent $\mathbf{Q}_\text{pose}$ from self-attention and image cues latent $\mathbf{Q}_\text{image}$ from deformable cross-attention.
Existing DETR-based human pose estimators~\cite{edpose,grouppose,tan2024diffusionregpose} (built on~\cite{DETR,deformable-detr}) directly sum $\mathbf{Q}_\text{pose}$ and $\mathbf{Q}_\text{image}$ before the feedforward network (FFN), a rigid design that leads to degraded performance under low-light conditions.

Following the analysis framework of~\citet{elhage2021mathematical,kim2025peri}, we evaluate the ratio of Frobenius Norm of $\mathbf{Q}_\text{image}$ and $\mathbf{Q}_\text{pose}$ in typical low-light conditions that one or more human joints are barely visible under low-light conditions, as shown in~\cref{fig:l2norm-vis}.
When both knees are nearly indistinguishable from the background, the ratio remains approximately $1.7$, comparable to that of the clearly visible keypoints.
In fact, across a set of well-lit images, this ratio remains stable with a mean around $1.68$.
This clearly demonstrates that rigid summation fails to reduce the contribution of unreliable image cues when keypoints are barely visible, leading to incorrect pose estimation for both knees as shown in~\cref{fig:l2norm-before}.

To address this issue, we introduce DCA, which adaptively fuses the pose prior latent $\mathbf{Q}_\text{pose}$ from self-attention and image cues latent $\mathbf{Q}_\text{image}$ from deformable cross-attention.
DCA first concatenates $\mathbf{Q}_\text{pose}$ and $\mathbf{Q}_\text{image}$ in channel dimension as $\mathbf{Q}_\text{cat}$,
\begin{equation}
\mathbf{Q}_\text{cat} = \text{Concat}(\mathbf{Q}_\text{pose}, \mathbf{Q}_\text{image}).
\end{equation}
Then, $\mathbf{Q}_\text{cat}$ is fed into a two-layer MLP that reduces the channel dimension to two.
A softmax function is then applied to produce two competitive weights corresponding to the pose-prior and image-cue features,
\begin{equation}
(\mathbf{w}_\text{pose}, \mathbf{w}_\text{image}) = \text{softmax}(\text{MLP} (\mathbf{Q}_\text{cat})).
\end{equation}
where $\mathbf{w}_\text{pose}$ and $\mathbf{w}_\text{image}$ are the obtained weights for $\mathbf{Q}_\text{pose}$ and $\mathbf{Q}_\text{image}$.
We then apply Hadamard product of the weights and their corresponding latent, where the new queries $\mathbf{Q}$ for the subsequent FFN and following layers are obtained by a weighted sum of $\mathbf{Q}_\text{pose}$ and $\mathbf{Q}_\text{image}$,
\begin{equation}
\mathbf{Q} = \mathbf{w}_\text{pose} \odot \mathbf{Q}_\text{pose} \oplus \mathbf{w}_\text{image} \odot \mathbf{Q}_\text{image}.
\end{equation}
Notably, DCA assigns different weights for different keypoints of a human instance, allowing the model to rely more on pose priors when a keypoint is less visible, and more on image cues when a keypoint is visible.
With only two additional linear layers, DCA enables the model to balance the influence of image cues and pose prior, leading to improved performance under low-light conditions as shown in~\cref{fig:l2norm-vis,fig:l2norm-after}.

\noindent\textbf{Pose Estimation Training}
%
For the human pose estimation model, we adopt the loss formulation from ED-Pose~\cite{edpose}. We employ a set-based Hungarian loss that forces
a unique prediction for each ground-truth box and keypoint. The total loss is a weighted sum of classification $\mathcal{L}_c$, human box regression $\mathcal{L}_h$, and keypoint regression loss $\mathcal{L}_k$. Notably, $\mathcal{L}_k$ simply
consists of the normal L1 loss and the constrained L1 loss named Object Keypoint Similarity (OKS) loss~\cite{rw-petr} without any dense supervision (e.g., heatmap). Please refer to the supplementary material for details.


\section{Experiment}
\label{sec:exp}
%

\noindent\textbf{Datasets}
We evaluate UDAPose on the ExLPose dataset~\cite{ExLPose_2023_CVPR}, specifically designed for benchmarking 2D human pose estimation in extremely low-light conditions.
%
ExLPose provides two distinct test sets: ExLPose-OCN and ExLPose-test.
ExLPose-OCN contains 360 real low-light images captured at night using A7M3 and RICOH3 cameras.
ExLPose-test consists of 491 optically filtered images, where brightness is reduced by a factor of 100.
ExLPose-test, also referred to as Low-Light All (LL-A), is further divided into three difficulty levels: Low-Light Normal (LL-N), Low-Light Hard (LL-H), and Low-Light Extreme (LL-E).

To validate our method's generalization ability, we performed cross-dataset evaluation on EHPT-XC~\cite{ehpt-xc}.
EHPT-XC is a novel hybrid dataset combining RGB and event data, specifically designed for human pose estimation and tracking in challenging low-light and motion blur conditions.
%
%
Given that some RGB data in EHPT-XC primarily exhibits motion blur without low-light conditions, we combined the train and test split of EHPT-XC and selected a specific subset of 12 scenes (1200 images) under low-light conditions for cross-dataset evaluation.
%

\begin{table}[!t]
\fontsize{9pt}{10pt}\selectfont 
\setlength{\tabcolsep}{1.2mm} 
\centering
    \begin{tabular}{l|r|rrrr}
    \hline
    \hline
    \multirow{2}{*}{\textbf{Methods}}            & \multicolumn{5}{c}{\textbf{AP$^{\uparrow}$@.50:.95}} \\ 
                                                                     \cline{2-6} 
                                                                     &  WL  & LL-A & LL-N & LL-H & LL-E  \\ 
    \hline
    RFormer~\cite{cai2023retinexformer}          & 60.0 & 4.5 & 15.2 & 0.3 & 0.8  \\
    DarkIR~\cite{feijoo2025darkir}               & 60.2 & 6.1 & 17.4 & 1.3 & 1.2 \\
    LightenDiff~\cite{jiang2024lightendiffusion} & 60.1 & 5.6 & 13.9 & 0.7 & 0.8  \\ 
    QuadPrior~\cite{quadprior}                   & 60.2 & 8.9 & 19.3 & 4.6 & 0.3  \\
    \hline
    CycleGAN~\cite{CycleGAN2017}                 & 61.3 & \underline{19.6} & \underline{33.7} & \underline{17.9} & 3.3  \\
    UNIT~\cite{UNIT}                         & 54.1 &  7.4 &  16.1 &  3.6 & 0.9  \\
    UNSB~\cite{kim2023unsb}                      & 57.5 & 15.8 & 28.3 & 13.8 & 1.8  \\
    EnCo~\cite{cai2024enco}                      & 60.0 & 17.2 & 31.9 & 16.2 & 2.9 \\
    UDA-HE~\cite{kim2022unified}                 & 53.4 & 13.2 & 22.4 & 12.7 & 1.8  \\
    ELLA~\cite{ella-eccv}           & \underline{61.5} & 18.6 & 32.3 & 17.2 &  \underline{3.4}  \\
    \hline
    \textbf{Ours}                                                             & \textbf{67.3} & \textbf{27.0} & \textbf{38.7} & \textbf{28.0} &  \textbf{11.7} \\ 
    \hline
    \hline
    \end{tabular}
\centering
\caption{
    Evaluation mAP on ExLPose-test comparing image enhancement and domain adaptation methods. 
    All methods are trained only on augmented images and well-lit annotations, without using low-light ground truth or paired data.
    %
    %
    The best is \textbf{bold}. 
    The second best is \underline{underlined}.
}
\label{tab:exlpose-test-AP}
\end{table}

\noindent\textbf{Evaluation metrics}
%
We evaluate our method following the COCO evaluation protocol~\cite{lin2014microsoft}, consistent with existing methods~\cite{ExLPose_2023_CVPR,ella-eccv, ehpt-xc}, on ExLPose-test, ExLPose-OCN, and EHPT-XC.
%
For each subset, we report Average Precision (AP) and Average Recall (AR) across multiple thresholds (@0.5:0.95) as the primary performance metrics.

\noindent\textbf{Implementation details}
We adopt SD-2.1-base as our backbone model for low-light data synthesis.
During training data synthesis, we use DDIM~\cite{song2020ddim} as our solver and 50 steps in sampling process.
%
%
We train LCIM using Adam optimizer~\cite{kingma2014adam} over 400
epochs in total, with an initial learning rate of $4 \times 10^{-6}$ that decreases
to $4 \times 10^{-7}$ after 300 epochs.
We train with a batch size of 32 across 4 NVIDIA RTX4090 GPUs and set the
weight $\lambda$ in~\cref{equ:llcd_loss_revised} to $4 \times 10^{-4}$ during training.
Our framework uses ED-Pose~\cite{edpose} with Swin-T~\cite{liu2021swin} pretrained on ImageNet 22k~\cite{deng2009imagenet} as the backbone for the human pose estimation model.
Training is performed with a batch size of 16 on 2 NVIDIA RTX PRO 6000 Blackwell GPUs.

\noindent\textbf{Baselines} For comparative evaluation, we benchmark against state-of-the-art methods in two categories: image enhancement (RFormer, DarkIR, QuadPrior, and LightenDiff~\citep{cai2023retinexformer,feijoo2025darkir,quadprior,jiang2024lightendiffusion}) and domain adaptation (CycleGAN, UNIT, UDA-HE, UNSB, EnCo~\citep{CycleGAN2017,UNIT,kim2022unified,kim2023unsb,cai2024enco}).
We also include ELLA~\cite{ella-eccv}, the state-of-the-art (SOTA) method for low-light human pose estimation.
To ensure a fair comparison, we use ED-Pose~\cite{edpose} as the backbone network across all methods, including our own.
While ELLA~\cite{ella-eccv} adopts a dual-teacher-student framework, where the student model distills knowledge from dual teachers utilizing low-light images, our method instead focuses on low-light synthesis and balancing pose priors and visual cues.
%
To evaluate the performance of data synthesis, the low-light data augmentation part of ELLA is applied in the comparison.
A full comparison to ELLA’s complete dual-teacher-student framework is provided in the supplementary material.
%

\subsection{Performance on ExLPose-test}
\begin{table}[!t]
\fontsize{9pt}{10pt}\selectfont 
\setlength{\tabcolsep}{1.2mm} 
\centering
    \begin{tabular}{l|r|rrrr}
    \hline
    \hline
    \multirow{2}{*}{\textbf{Methods}}            & \multicolumn{5}{c}{\textbf{AR$^{\uparrow}$@.50:.95}} \\ 
                                                                     \cline{2-6} 
                                                                     &  WL  & LL-A & LL-N & LL-H & LL-E  \\ 
    \hline
    RFormer~\cite{cai2023retinexformer}          & 71.5 &  9.7 & 25.4 & 4.1 & 0.6  \\
    DarkIR~\cite{feijoo2025darkir}               & 72.0 & 11.1 & 29.3 & 7.3 & 0.8 \\
    LightenDiff~\cite{jiang2024lightendiffusion} & 71.8 & 10.2 & 22.9 & 4.9 & 0.5  \\ 
    QuadPrior~\cite{quadprior}                   & 71.9 & 15.5 & 30.9 & 11.7 & 1.0  \\
    \hline
    CycleGAN~\cite{CycleGAN2017}                 & 72.1 & 28.7 & \underline{45.6} & 27.4 &  8.9  \\
    UNIT~\cite{UNIT}                         & 68.0 &  14.2 &  26.7 &  10.3 &  3.3  \\
    UNSB~\cite{kim2023unsb}                      & 69.4 & 23.8 & 39.7 &  22.5 &  5.7  \\
    EnCo~\cite{cai2024enco}                      & 71.8 & 27.4 & 41.5 & 26.3 & 5.7 \\
    UDA-HE~\cite{kim2022unified}                 & 67.2 & 21.4 & 32.7 & 20.9 & 5.8  \\
    ELLA~\cite{ella-eccv}           & \underline{72.7} & \underline{28.9} & 45.0 & \underline{28.0} &  \underline{9.7}  \\
    \hline
    \textbf{Ours}                                                             & \textbf{75.0} & \textbf{36.5} & \textbf{48.2} & \textbf{37.4} &  \textbf{20.4} \\ 
    \hline
    \hline
    \end{tabular}
\centering
\caption{
    Evaluation mAR on ExLPose-test comparing image enhancement and domain adaptation methods, following~\cref{tab:exlpose-test-AP}.
    %
    %
    The best is \textbf{bold}. 
    The second best is \underline{underlined}.
}
\label{tab:exlpose-test-AR}
\end{table}
As shown in~\cref{tab:exlpose-test-AP,tab:exlpose-test-AR}, UDAPose consistently outperforms all baselines on the ExLPose-test set. On average over all low-light conditions (LL-A), UDAPose achieves 27.0 AP and 36.5 AR, surpassing the best-performing baseline, CycleGAN~\cite{CycleGAN2017} and ELLA~\cite{ella-eccv}, by 7.4 AP (a 37.8\% relative improvement) and 7.6 AR (a 26.3\% relative improvement). 
The performance gap widens as lighting conditions deteriorate. Specifically, UDAPose leads by 5.0 AP and 2.6 AR on the normal subset (LL-N) and by 10.1 AP (a 56.4\% relative gain) and 9.4 AR (a 33.6\% relative gain) on the hard subset (LL-H).
The advantage is clearest on the extreme subset (LL-E), where UDAPose delivers a more than three-fold improvement over ELLA on AP (11.7 vs. 3.4 AP) and more than double on AR (20.4 vs. 9.7 AR). 

\subsubsection{Qualitative Analysis}
\begin{figure*}[t!]
{  \centering

    \begin{minipage}[c]{0.196\linewidth}
		\includegraphics[width=\linewidth]{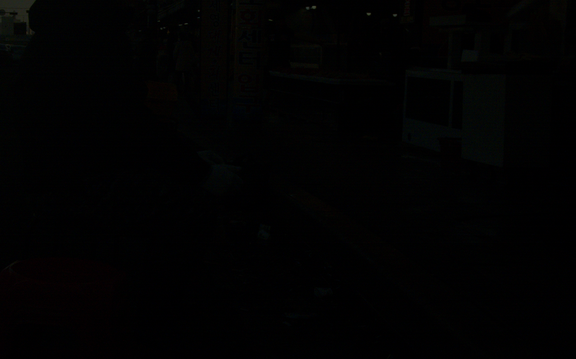}
	\end{minipage}
    \begin{minipage}[c]{0.196\linewidth}
        \includegraphics[width=\linewidth]
        {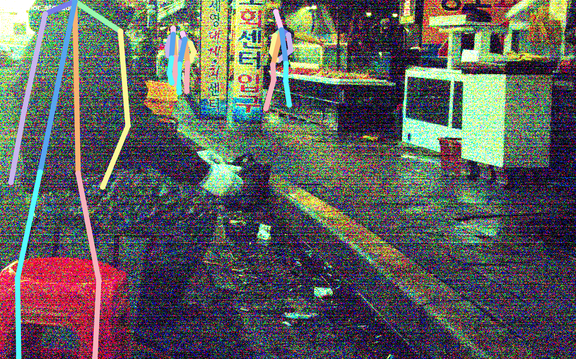}
	\end{minipage}
	\begin{minipage}[c]{0.196\linewidth}
        \includegraphics[width=\linewidth]
        {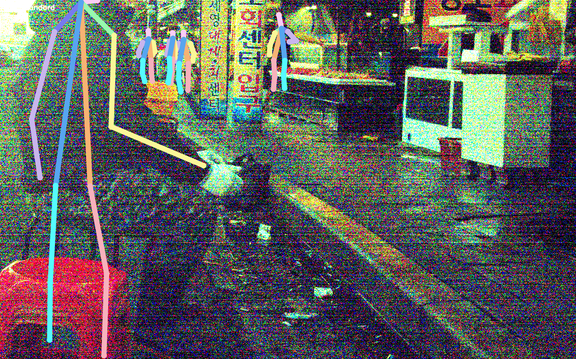}
	\end{minipage}
    \begin{minipage}[c]{0.196\linewidth}
        \includegraphics[width=\linewidth]
        {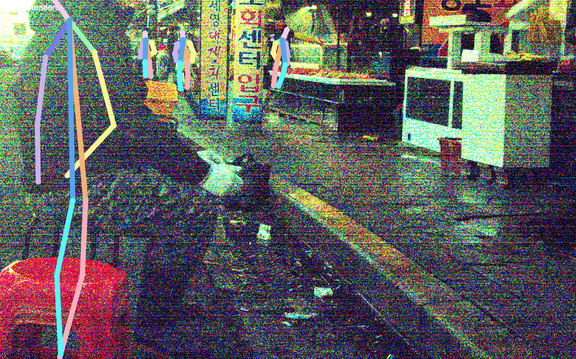}
	\end{minipage}
	\begin{minipage}[c]{0.196\linewidth}
        \includegraphics[width=\linewidth]
        {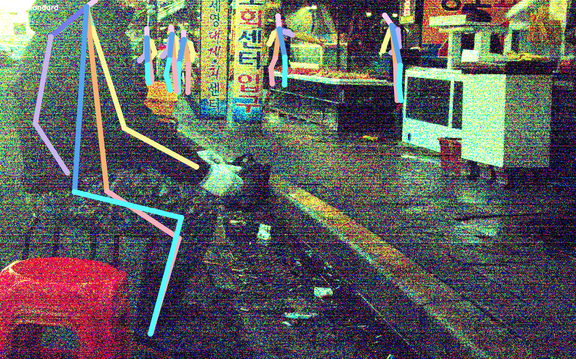}
	\end{minipage}

    \begin{minipage}[c]{0.196\linewidth}
		\includegraphics[width=\linewidth]{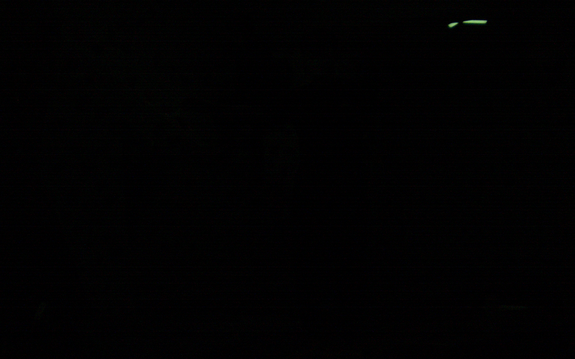}
	\end{minipage}
    \begin{minipage}[c]{0.196\linewidth}
        \includegraphics[width=\linewidth]
        {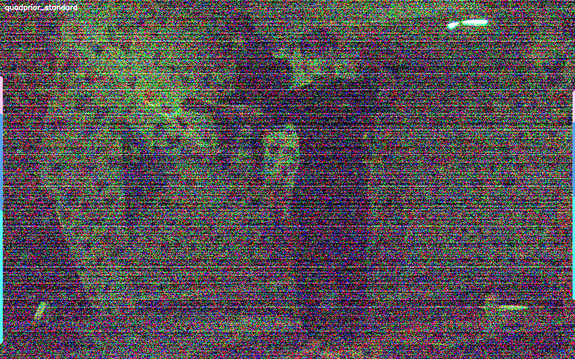}
	\end{minipage}
	\begin{minipage}[c]{0.196\linewidth}
        \includegraphics[width=\linewidth]
        {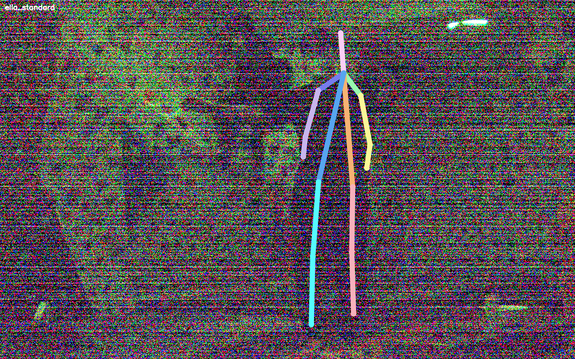}
	\end{minipage}
    \begin{minipage}[c]{0.196\linewidth}
        \includegraphics[width=\linewidth]
        {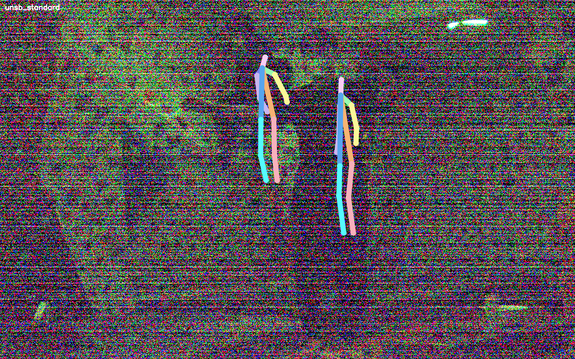}
	\end{minipage}
	\begin{minipage}[c]{0.196\linewidth}
        \includegraphics[width=\linewidth]
        {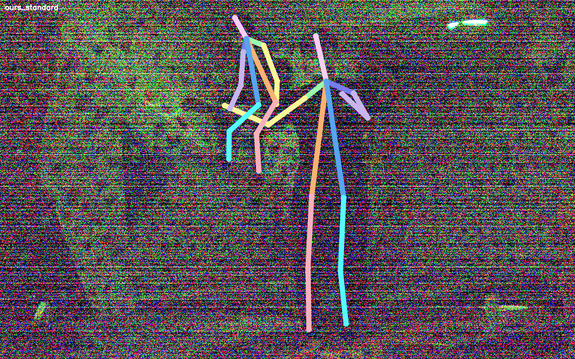}
	\end{minipage}
 
    \begin{minipage}[c]{0.196\linewidth}
		\centerline{\scriptsize{Input Image}}
	\end{minipage}
    \begin{minipage}[c]{0.196\linewidth}
		\centerline{\scriptsize{QuadPrior~\cite{quadprior}}}
	\end{minipage}
	\begin{minipage}[c]{0.196\linewidth}
		\centerline{\scriptsize{ELLA~\cite{ella-eccv}}}
	\end{minipage}	
    \begin{minipage}[c]{0.196\linewidth}
		\centerline{\scriptsize{UNSB~\cite{kim2023unsb}}}
	\end{minipage}
    \begin{minipage}[c]{0.196\linewidth}
		\centerline{\scriptsize{Ours}}
	\end{minipage}
\captionof{figure}{
Qualitative comparisons of our method with existing baselines, including image enhancement~\cite{quadprior}, domain adaptation~\cite{ella-eccv}, and image translation method~\cite{kim2023unsb}. From top to bottom, we show samples from low-light normal, hard, and extreme sets from ExLPose-test~\cite{ExLPose_2023_CVPR}. Our approach consistently outperforms the existing methods in human pose estimation across all scenarios.
Images are scaled up for visualization purpose only.}
\label{fig:qualitative}
}
\end{figure*}

\cref{fig:qualitative} provides a qualitative comparison against representative methods. In normal low-light conditions, UDAPose already yields more precise joint localization. As lighting degrades, our superiority becomes more evident. 
In hard low-light, competing methods produce poses with significant limb errors, whereas UDAPose maintains an accurate skeletal structure. Under extreme low-light, where the subject is barely visible, other methods generate fragmented and anatomically implausible poses. In contrast, UDAPose successfully reconstructs a complete and coherent human pose. These visual results support our quantitative findings, confirming that our synthetic data generation and adaptive fusion of learned pose priors and image cues enable robust pose estimation in challenging low-light scenarios.

\subsection{Performance on ExLPose-OCN}
\begin{table}[!t]
\fontsize{9pt}{10pt}\selectfont 
\setlength{\tabcolsep}{1.2mm} 
\centering
    \begin{tabular}{l|r|rr|r|rr}
    \hline
    \hline
    \multirow{2}{*}{\textbf{Methods}} & \multicolumn{3}{c|}{\textbf{AP$^{\uparrow}$@.50:.95}} & \multicolumn{3}{c}{\textbf{AR$^{\uparrow}$@.50:.95}}   \\ 
                                                                     \cline{2-7} 
                & Avg. & \makecell{A7 \\ M3} & \makecell{RIC \\ OH3} & Avg. & \makecell{A7 \\ M3} & \makecell{RIC \\ OH3}  \\ 
    \hline
    RFormer~\cite{cai2023retinexformer}          & 27.5 & 29.4 &  25.7 & 43.7 & 47.2 & 40.3 \\
    DarkIR~\cite{feijoo2025darkir}               & 28.9 & 30.3 & 27.6 & 47.0 & 48.7 & 45.4 \\
    LightenDiff~\cite{jiang2024lightendiffusion} &  25.3 &  29.1 &   21.5 & 40.6 & 45.4 & 35.9 \\ 
    QuadPrior~\cite{quadprior}                  & 29.3 & 30.6 &  28.0  & 48.8 & 50.4 & 47.2\\
    \hline
    CycleGAN~\cite{CycleGAN2017}                & 45.1 & 47.5 &  42.7  & 60.7 & 63.7 & 57.8  \\
    UNIT~\cite{UNIT}                         & 35.1 & 40.7 &  29.6 & 55.2 & 61.9 & 48.6 \\
    UNSB~\cite{kim2023unsb}                      & 41.6 & 43.5 &  39.8 & 59.7 & 60.9 & 58.5 \\
    EnCo~\cite{cai2024enco}                      & 43.2 & 45.4 & 41.1 & 58.4 & 61.5 & 55.4 \\
    UDA-HE~\cite{kim2022unified}                &  39.2 &  42.4 &   36.0 & 57.1 & 61.8 & 52.4 \\
    ELLA~\cite{ella-eccv}           & \underline{46.0} & \underline{48.5} &  \underline{43.5} & \underline{62.3} & \underline{65.5} & \underline{59.1} \\
    \hline
    \textbf{Ours}                                                             & \textbf{51.4} & \textbf{55.0} &  \textbf{47.9} & \textbf{65.1} & \textbf{68.1} & \textbf{62.2}  \\
    \hline
    \hline
    \end{tabular}
\centering
\caption{
    Evaluation on ExLPose-OCN, following identical setup as in \cref{tab:exlpose-test-AP}. 
    %
    %
    %
    The best is \textbf{bold}. 
    The second best is \underline{underlined}.
    %
}
\label{tab:exlpose-ocn}
\end{table}
\cref{tab:exlpose-ocn} presents our quantitative results on ExLPose-OCN. Our model achieves an average AP of 51.4 and an average AR of 65.1, a 5.4 AP improvement (11.7\% relative gain) and a 2.8 AR improvement (4.5\% relative gain) over the previous state-of-the-art~\cite{ella-eccv}. 
%
These results show that our approach enables robust pose estimation in real-world low-light conditions where annotated data is scarce.

\begin{table}[!t]
\fontsize{9pt}{10pt}\selectfont 
\setlength{\tabcolsep}{1mm} 
\centering
\begin{tabular}{l|rrr|rrr}
    \hline\hline
    \multirow{2}{*}{\textbf{Methods}} & \multicolumn{3}{c|}{\textbf{AP$^{\uparrow}$}} & \multicolumn{3}{c}{\textbf{AR$^{\uparrow}$}} \\
    \cline{2-7}
    & {\scriptsize@.50:.95} & {\scriptsize@.50} & {\scriptsize@.75} & {\scriptsize@.50:.95} & {\scriptsize@.50} & {\scriptsize@.75} \\
    \hline
    RFormer~\cite{cai2023retinexformer} & 8.8 & 20.3 & 7.9 & 21.3 & 39.2 & 18.9  \\
    DarkIR~\cite{feijoo2025darkir}      & 12.5 & 25.2 & 11.3 & 28.9 & 48.5 & 27.1 \\
    LightenDiff~\cite{jiang2024lightendiffusion} & 9.7 & 17.7 & 8.3 & 22.0 & 38.4 & 20.3  \\
    QuadPrior~\cite{quadprior}   & 12.9 & 24.5 & 11.5 & 28.1 & 47.7 & 27.5  \\
    \hline
    CycleGAN~\cite{CycleGAN2017} & 20.7 & 36.9 & 19.2 & 45.0 & 69.6 & 45.5  \\
    UNIT~\cite{UNIT} & 11.2 & 24.5 &  8.7 & 35.5 & 64.2 & 32.5  \\
    UNSB~\cite{kim2023unsb}  & 16.9 & 30.1 &  15.7 & 40.5 & 67.1 & 39.2  \\
    EnCo~\cite{cai2024enco}  & 18.3 & 33.4 & 17.1 & 42.3 & 68.1 & 42.8 \\
    UDA-HE~\cite{kim2022unified}  & 15.4 &  29.6 & 14.8 &  38.7 &  66.7 & 37.8  \\
    ELLA~\cite{ella-eccv} & \underline{23.6} & \underline{38.3} & \underline{22.7} & \underline{48.2} & \underline{73.2} & \underline{48.6}   \\
    \hline
    \textbf{Ours} & \textbf{31.0} & \textbf{51.1} & \textbf{30.3} & \textbf{51.3} & \textbf{76.4} & \textbf{53.8}  \\
    \hline\hline
\end{tabular}
\centering
\caption{
    Cross-dataset validation on EHPT-XC~\cite{ehpt-xc}, using the model weights as in \cref{tab:exlpose-test-AP}.
    Best is \textbf{bold}, second best \underline{underlined}.
}
\label{tab:zero-shot}
\end{table}

\begin{table}[!t]
\fontsize{9pt}{10pt}\selectfont 
\setlength{\tabcolsep}{1.2mm} 
\centering
    \begin{tabular}{l| r | rrr | rr | r}
    \hline
    \hline
          & \multicolumn{7}{c}{\textbf{AP$^{\uparrow}$@{0.5:0.95}}} \\
    \cline{2-8}         &  WL  & LL-N & LL-H & LL-E & \makecell{A7 \\ M3} & \makecell{RIC \\ OH3} & \makecell{EHPT \\ -XC} \\
    \hline
    Well-lit & 60.1 &    3.4 &  0.4 &  0.2 &  11.3 &   13.4 & 0.3 \\
    HM  & 60.0 &   21.3 &  1.3 &  0.5 &  15.1 &   11.4 & 1.5 \\
    \hline
    Baseline SD  & 60.0 & 23.7 & 7.2 & 0.0 & 30.1 & 25.4 & 3.2 \\
    + AIN     & 60.1  & 25.2 & 8.8 & 2.4 & 33.0 & 27.8 & 6.1 \\
    + LCIM    & 60.1  & 31.5 & 20.7 & 7.8 & 43.1 & 39.8 & 19.5 \\
    + DHF           & 60.2  & 35.3 & 25.3 & 9.4 & 48.9 & 45.6 & 24.4 \\
    + DCA & \textbf{67.3}  & \textbf{38.7} & \textbf{28.0} & \textbf{11.7} & \textbf{55.0} & \textbf{47.9} & \textbf{31.0} \\
    \hline
    \hline
  \end{tabular}
  \centering
  \caption{
  Ablation study of our proposed modules on ExLPose-test, ExLPose-OCN, and EHPT-XC. 
  Well-lit: pose model trained with well-lit images only.
  HM: pose model adapted with synthetic low-light images using histogram matching.
  AIN is a normalization step (see supplementary).
  The best is \textbf{bold}. 
  }
  \label{tab:abl}
\end{table}

\subsection{Cross-dataset validation on EHPT-XC}
To assess our model's generalization ability, we perform a cross-dataset validation on the EHPT-XC dataset~\cite{ehpt-xc}, which features challenging real-world conditions such as motion blur and low light.
As shown in~\cref{tab:zero-shot}, our method consistently outperforms all state-of-the-art baselines across all Average Precision (AP) and Average Recall (AR) metrics.
Specifically, our approach achieves an AP@0.5:0.95 of 31.0, surpassing the strongest baseline, ELLA~\cite{ella-eccv}, by 7.4 points.
This strong performance also extends to other metrics, highlighting our model’s ability to generalize to unseen and degraded data without fine-tuning.


\subsection{Ablation Studies}

\subsubsection{Baselines}
We first establish two baselines. A standard pose estimator trained only on well-lit data fails on low-light images (e.g., 3.4 AP on LL-N), highlighting a significant domain gap. 
Adapting this model with a simple histogram matching technique offers only a marginal improvement (21.3 AP on LL-N), indicating that basic color transformations are insufficient. 
A second baseline, ``Baseline SD'', following~\citet{chung2024style}, achieves decent performance by synthesizing low-light images (23.7 AP on LL-N) but still struggles with more challenging conditions (7.2 AP on LL-H).

\subsubsection{Component Analysis}
As shown in~\cref{tab:abl}, building on ``Baseline SD'', we first integrate LCIM, which yields the most substantial performance leap. The LCIM dramatically improves results on the most difficult subsets, increasing the AP on LL-H from 7.2 to 20.7 and on LL-E from 0.0 to 7.8.
This highlights the importance of LCIM in transferring low-light features from the reference image.
In addition, the DHF module improves performance by modeling frequency-domain attributes, delivering notable performance gains on challenging data (LL-H: +4.6 AP, LL-E: +1.6 AP).
Lastly, DCA is added to improve our pose model's capability to deal with low-visibility conditions, leading to consistent improvement across all subsets, from LL-N (+3.4 AP) to more challenging LL-H (+2.7 AP) and LL-E (+2.3 AP).
Overall, each component plays a clear role, and together they improve the model’s ability to handle varied low-light scenarios.

\subsection{Scaling to Larger Well-lit Source Data}
Our framework can use any well-lit pose dataset as the source for low-light synthesis, since it only requires well-lit images with pose annotations and unlabeled low-light reference images.
To evaluate scaling potential, we replace the ExLPose well-lit set ($\sim$2k images) with CrowdPose~\cite{li2019crowdpose} ($\sim$12k images, approximately 6$\times$ larger) as the source for our synthesis pipeline.
The low-light reference images and test protocol remain unchanged.

As shown in~\cref{tab:crowdpose}, using CrowdPose as the source improves performance across all evaluation sets.
On ExLPose-test, the model improves by 9.2 AP on both LL-N and LL-H, and by 7.7 AP on LL-E.
On the cross-dataset EHPT-XC benchmark, the gain reaches 14.7 AP (31.0 $\rightarrow$ 45.7).
The largest improvements appear on ExLPose-OCN, where A7M3 improves by 15.5 AP (55.0 $\rightarrow$ 70.5) and RICOH3 by 21.8 AP (47.9 $\rightarrow$ 69.7).
Notably, the CrowdPose variant's performance on ExLPose-OCN (A7M3: 70.5, RICOH3: 69.7) approaches its own well-lit performance (WL: 71.7), showcasing that our synthesis pipeline substantially reduces the domain gap when given sufficient well-lit source data.

\begin{table}[t]
\fontsize{9pt}{10pt}\selectfont 
\setlength{\tabcolsep}{0.8mm} 
\centering
    \begin{tabular}{l| c | rrr | cc | c}
    \hline
    \hline
          & \multicolumn{7}{c}{\textbf{AP$^{\uparrow}$@{0.5:0.95}}} \\
    \cline{2-8}            &  WL  & LL-N & LL-H & LL-E & \makecell{A7 \\ M3} & \makecell{RIC \\ OH3} & \makecell{EHPT \\ -XC} \\
    \hline
    Ours (ExLPose)         & 67.3 & 38.7 & 28.0 & 11.7 & 55.0 & 47.9 & 31.0  \\
    Ours (CrowdPose)       & 71.7 & 47.9 & 37.2 & 19.4 & 70.5 & 69.7 & 45.7  \\

    \hline
    \hline
  \end{tabular}
  \centering
  \caption{
    Comparison results of using ExLPose well-lit and CrowdPose to construct synthetic low-light training data.
  }
  \label{tab:crowdpose}
\end{table}

\section{Conclusion}
\label{sec:con}
In this work, we introduced UDAPose, a novel domain-adaptive framework for human pose estimation under low-light conditions.
UDAPose leverages SD to synthesize low-light images, incorporating the DHF with LCIM to retain key low-light details.
UDAPose also integrates DCA to balance unreliable visual cues under poorly illuminated conditions with learned pose priors. 
This results in improved performance under extreme lighting conditions.
Our approach outperforms both rule-based and learning-based augmentation methods, achieving substantial performance gains on ExLPose and EHPT-XC.
These results demonstrate UDAPose’s effectiveness in addressing limitations of existing low-light human pose estimation methods and improving human pose estimation performance in real-world low-visibility scenarios.

\section{Acknowledgment}
\label{sec:ack}
This work was supported in part by the Mississippi Impact Grant (MIG), Office for Research and Economic Development, University of Mississippi.
We thank the anonymous reviewers and the area chair for their constructive feedback, which helped improve this paper.



{
    \small
    \bibliographystyle{ieeenat_fullname}
    \bibliography{main}
}

\clearpage
\setcounter{page}{1}
\maketitlesupplementary

\section{Implementation and Experimental Details}

\subsection{Human Pose Estimation Model}
%
\begin{figure}[!t]
  \centering
  \includegraphics[width=1.0\columnwidth]{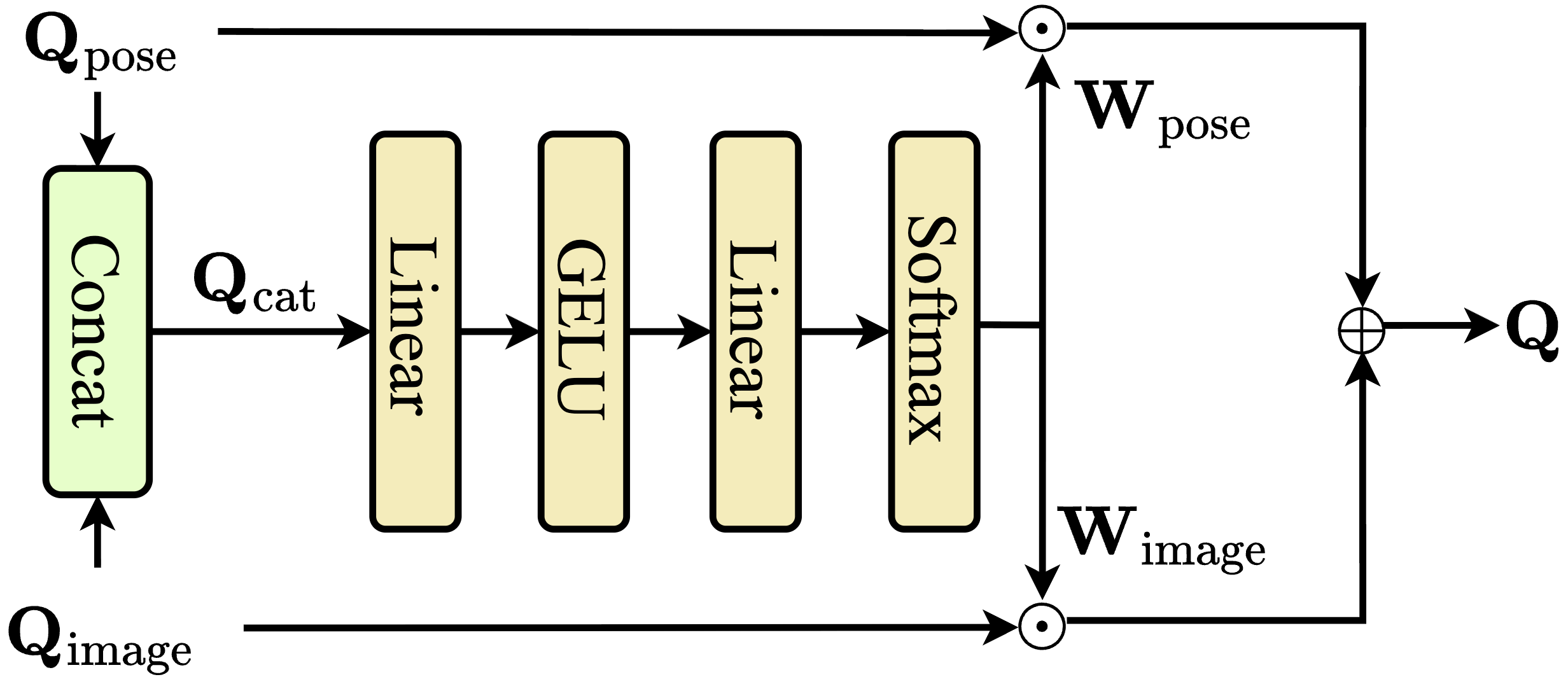}
  \caption{
    The architecture of our DCA module.
  }
  \label{fig:dca_arch}
\end{figure}
\subsubsection{Architecture of DCA}
We present the details of the proposed Dynamic Control of Attention (DCA) module in~\cref{fig:dca_arch}.
As described in the main paper (Sec.~3 Method), DCA first concatenates $\mathbf{Q}_\text{pose}$ with $\mathbf{Q}_\text{image}$ as $\mathbf{Q}_\text{cat}$.
Then $\mathbf{Q}_\text{cat}$ goes through two-layer MLP ended with Softmax to acquire $\mathbf{W}_\text{pose}$ and $\mathbf{W}_\text{image}$, adaptive weights for pose priors $\mathbf{Q}_\text{pose}$ and image cues $\mathbf{Q}_\text{image}$, respectively.
At last, DCA fuses weighted sum of pose prior $\mathbf{Q}_\text{pose}$ and image cues $\mathbf{Q}_\text{image}$ as output $\mathbf{Q}$ for subsequent FFN and following decoder layers. 
As shown in Fig.~3 of the main paper, DCA is placed after deformable cross-attention within each decoder layer to substitute original direct sum of residual connection.

\subsubsection{Loss Functions}
The overall loss functions of human pose estimation model can be formulated as:
\begin{equation}
    \mathcal{L} = \mathcal{L}_h + \mathcal{L}_c + \mathcal{L}_k
\end{equation}
\begin{equation}
    \mathcal{L}_h = \mu \left| H- \hat{H} \right| + \beta (1-\text{GIoU})
\end{equation}
\begin{equation}
\begin{split}
    \mathcal{L}_c &= -\lambda\alpha(1-p_t)^\gamma\log(p_t), \\ &\text{where}~p_t = p~\text{if}~y=1, p_t=1-p~\text{if}~y\neq1
\end{split}
\end{equation}
\begin{equation}
\begin{split}
    \mathcal{L}_k &= \omega \left | P-\hat{P} \right | \\ &+ \theta \frac{\sum_i^K \exp{\left(- \left | P_i-\hat{P_i} \right | / 2s^2k_i^2  \right) \delta (v_i>0)}}{\sum^K_i \delta (v_i>0)}
\end{split}
\end{equation}
where $\mathcal{L}_h$ is for human box regression that contains L1 loss and GIoU~\cite{rezatofighi2019generalized} loss, $\mathcal{L}_c$ is for human classification, which is a focal loss~\cite{lin2017focal} with $\alpha = 0.25, \gamma = 2$, and $\mathcal{L}_k$ is for keypoint regression that includes L1 loss and the constrained L1 loss-OKS loss~\cite{rw-petr}.
$| H- \hat{H} |$ is the L1 distance between the predicted human boxes and the ground-truth ones.
$y \in \pm1$ specifies the ground-truth class, and $p \in [0,1]$ is the estimated probability for the class with label $y=1$.
$| P-\hat{P} |$ is the L1 distance between predicted keypoints inside a human and the ground-truth ones.
$| P_i-\hat{P_i} |$ is the L1 distance between the $i$-th predicted keypoint and ground-truth one, $v_i$ is the visibility flag of the ground truth, $s$ is the object scale, and $k_i$ is the per-keypoint constant that controls falloff.
The loss coefficients $\mu, \beta, \lambda, \omega, \theta$ are 5, 2, 2, 10, 4.

\subsection{Datasets}
As introduced in the main paper, we evaluate UDAPose on the ExLPose dataset~\cite{ExLPose_2023_CVPR}, which is specifically for benchmarking 2D human pose estimation in extremely low-light conditions.
The ExLPose training set consists of 2,065 well-lit and optically filtered low-light image pairs, with pose annotations following the CrowdPose format~\cite{li2019crowdpose}.
These images span 251 indoor and outdoor scenes, with low-light versions generated using a dual-camera system under varying conditions to simulate diverse low-light scenarios. 
ExLPose provides two test sets: ExLPose-test and ExLPose-OCN.
ExLPose-test, also referred to as Low-Light All (LL-A), is further divided into three difficulty levels: Low-Light Normal (LL-N), Low-Light Hard (LL-H), and Low-Light Extreme (LL-E).
To validate our method's generalization ability, we also performed cross-dataset validation on EHPT-XC~\cite{ehpt-xc}.
EHPT-XC is a dataset combining RGB and event camera data for human pose estimation and tracking in challenging low-light and motion blur conditions.
It encompasses RGB video frames from 158 diverse sequences, along with pixel-wise aligned and temporally synchronized event streams, and annotations containing 38K 2D keypoints and bounding boxes with track IDs.
To focus on low-light conditions, we combined the train and test split of EHPT-XC and constructed a specific subset of 12 scenes (1200 images) for cross-dataset validation.

\subsection{Discussion on Dual-Camera Data Usage}

The dual-camera setup in ExLPose~\cite{ExLPose_2023_CVPR} is an effective design that enables paired data collection by transferring annotations from well-lit images to their low-light counterparts. 
However, this setup relies on hardware-specific acquisition and cannot be applied to synthesize low-light images from existing well-lit human pose datasets.
In addition, it is not easily scalable for collecting new data, as it requires a specialized camera system rather than standard cameras. 
In contrast, our method allows leveraging existing well-lit human pose datasets with available annotations, enabling flexible and scalable low-light data generation without requiring specialized camera systems.

In our experiment, we use the dual-camera low-light images from ExLPose as style references. While such images may not fully represent real-world low-light conditions for supervised learning (e.g., due to optical filtering), they still capture useful characteristics such as illumination patterns and noise distributions. This usage is consistent with our goal of synthesizing low-light images from well-lit data, rather than directly training on limited low-light datasets.
Another reason is to ensure fair comparison with prior work~\cite{ella-eccv,ExLPose_2023_CVPR}. We use the same dataset as the source of style references, avoiding performance gains from larger or more diverse real nighttime datasets. Otherwise, improvements could be attributed to data scale instead of the proposed method. By using the same dataset, we isolate performance gains to the proposed method.

\begin{table}[t]
\fontsize{9pt}{10pt}\selectfont 
\setlength{\tabcolsep}{1.2mm} 
\centering
    \begin{tabular}{r| r | rrr | cc | c}
    \hline
    \hline
         & \multicolumn{7}{c}{\textbf{AP$^{\uparrow}$@{0.5:0.95}}} \\
    \cline{2-8}       &  WL  & LL-N & LL-H & LL-E & \makecell{A7 \\ M3} & \makecell{RIC \\ OH3} & \makecell{EHPT \\ -XC} \\
    \hline
    4,000             & 66.1 & 34.7 & 22.4 & 5.4 & 50.0 & 45.1 & 25.4  \\
    8,000             & 66.3 & 35.4 & 24.8 & 7.8 & 53.2 & 46.8 & 27.9  \\
    12,000            & 66.9 & 36.6 & 26.2 & 10.4 & 54.7 & 47.2 & 29.1  \\
    16,000            & 66.8 & 37.5 & 27.3 & 11.3 & 55.0 & 47.8 & 30.5  \\
    20,000            & 67.3 & 38.7 & 28.0 & 11.7 & 55.0 & 47.9 & 31.0 \\

    \hline
    \hline
  \end{tabular}
  \centering
  \vspace{-0.3cm}
  \caption{
    Performance~vs.~amount of synthetic training data.
  }
  \label{tab:scale}
\end{table}

\subsection{Implementation Details}

For each well-lit image in ExLPose, we randomly select one low-light image from ExLPose or ExLPose-OCN as its style reference, and repeat this process 10 times, yielding 20k synthetic low-light images for training. 
We further analyze data scaling in \cref{tab:scale}, where performance improves with more synthetic training images (with larger gains below 12k), and use 20k images in the main experiments. The synthesis cost is approximately 263 ms per image on an RTX 4090.


Following the pipeline of ED-Pose~\cite{edpose}, we adopt the overall pose-estimation framework and focus our contributions on the proposed DCA module. 
We utilize Swin-Transformer~\cite{liu2021swin} (Swin-T) pretrained on ImageNet-22k~\cite{deng2009imagenet} as the multi-scale image feature extraction backbone.
During training, we apply data augmentations including random crop, random flip, and random resize (shorter side in [480, 800], longer side $\leq$ 1333),  following DETR~\cite{DETR} and PETR~\cite{rw-petr}.
To accelerate the early-stage training, we adopt the human query denoising training strategy from DN-DETR~\cite{li2022dn-detr}.
We use the AdamW~\cite{kingma2014adam,loshchilov2017AdamW} optimizer with weight decay of $1 \times 10^{-4}$ and train our pose model on 2 NVIDIA RTX PRO 6000 GPUs with batch size 16 for 120 epochs on ExLPose~\cite{ExLPose_2023_CVPR}.
The initial learning rate is $1 \times 10^{-4}$ and is decayed at the 100th epoch by a factor of 0.1.
The channel dimension of the Transformer layers is set to 256. 
At test time, we resize each input image so that its shorter side is 800 pixels while keeping the longer side no more than 1333 pixels.
DCA introduces only 4.1\% inference overhead to pose model (39 ms vs. 37.4 ms per image) on an RTX PRO 6000.

\subsection{Experiment Settings}
For image enhancement methods~\cite{cai2023retinexformer,feijoo2025darkir,quadprior,jiang2024lightendiffusion}, we directly use the official checkpoints released by the authors to ensure a fair comparison.
When applying the image enhancement models for human pose estimation evaluation, we first convert low-light images from ExLPose-test, ExLPose-OCN and EHPT-XC using their models.
After that, we apply human pose estimation model trained on ExLPose well-lit images on these enhanced images to test performance.

For domain adaptive methods~\cite{CycleGAN2017,UNIT,kim2023unsb,cai2024enco,kim2022unified,ella-eccv}, we follow the same procedure as used in existing methods~\cite{ella-eccv}.
The human pose estimation model is first trained on ExLPose well-lit dataset and then finetuned on augmented low-light images.
At test time, we directly input low-light images from ExLPose-test, ExLPose-OCN and EHPT-XC to test their performance.

\section{Anatomical Consistency}
\begin{table}[!t]
\fontsize{9pt}{10pt}\selectfont 
\setlength{\tabcolsep}{1.1mm} 
\centering
    \begin{tabular}{l|cc|cc|c}
    \hline
    \hline
    \textbf{Methods}  & \textbf{PSNR}$^{\uparrow}$ & \textbf{SSIM}$^{\uparrow}$ & \textbf{LPIPS}$^{\downarrow}$ & \textbf{FID}$^{\downarrow}$ & \textbf{KL}$^{\downarrow}$ \\
    \hline
    CycleGAN~\cite{CycleGAN2017} &     36.56    &     0.76     &      0.26     &  50.20  &   0.028       \\
    UNIT~\cite{UNIT}          &     33.90    &     0.66     &      0.29     &  45.70  &   0.104       \\
    UNSB~\cite{kim2023unsb}      &     34.19    &     0.74     &      0.30     &  96.42  &   0.062       \\
    EnCo~\cite{cai2024enco}          & 35.99 & 0.75 & 0.28 & 48.71 & 0.031 \\
    \hline
    \textbf{Ours}                                             &\textbf{41.13}& \textbf{0.91}& \textbf{0.20} & \textbf{11.17} &\textbf{0.008} \\
    \hline
    \hline
    \end{tabular}
\centering
\caption{
    Evaluation for human pose anatomical consistency of our method and learning-based baselines.
}
\label{tab:consistency}
\end{table}
We evaluate the anatomical consistency of our generated low-light images, an important factor for reusing human pose annotations from well-lit datasets.
This evaluation also helps verify that our method preserves structural consistency and avoids unintended structure leakage from the style reference.
To this end, we synthesize low-light images from the well-lit inputs in ExLPose and evaluate them against the paired low-light images (i.e., ExLPose includes paired well-lit and low-light images.) using a comprehensive set of metrics.
We assess image fidelity at the pixel level with peak signal-to-noise ratio (PSNR) and structural similarity index measure (SSIM), and at the feature level with learned perceptual image patch similarity (LPIPS) and Fr\'echet Inception Distance (FID).
More importantly, to directly quantify anatomical integrity, we compute the Kullback–Leibler (KL) divergence between predicted heatmaps on our synthetic low-light images and on the paired low-light images from ExLPose. A pose estimator (DEKR~\cite{DEKR2021}) trained on the low-light data from ExLPose is used to predict the heatmaps in this experiment.

Learning-based adaptation methods (e.g., unpaired image-to-image translation or style transfer) with competitive performance are used here as baselines. As shown in~\cref{tab:consistency}, our method significantly outperforms across every metric.
Our method achieves a PSNR of 41.13 and an SSIM of 0.91, indicating superior pixel-level accuracy.
Furthermore, our method obtains the lowest LPIPS (0.20) and a remarkably low FID of 11.17, confirming that the generated images have higher perceptual quality and a feature distribution much closer to that of real images.
Importantly, our method obtains a KL divergence of only 0.008, which is much lower than the best-performing baseline (CycleGAN: 0.028). 
These results provide solid evidence that our generation process preserves the underlying human anatomical structure faithfully, which facilitates downstream human pose estimation tasks using our synthetic data.

\section{Comparison with ELLA and Supervised Low-Light Training}
\label{sec:comp_ella}
\begin{table}[!t]
\fontsize{9pt}{10pt}\selectfont 
\setlength{\tabcolsep}{1.2mm} 
\centering
    \begin{tabular}{l| r | rrr | rr}
    \hline
    \hline
          & \multicolumn{6}{c}{\textbf{AP$^{\uparrow}$@{0.5:0.95}}} \\
    \cline{2-7}                 &  WL  & LL-N & LL-H & LL-E & \makecell{A7 \\ M3} & \makecell{RIC \\ OH3} \\
    \hline
    Main (ELLA)             & \textbf{62.1} & 29.4 & 13.6 &  1.6 & 35.0 & 27.2  \\
    Main (Ours)             & 61.5 & 32.3 & 23.2 &  8.3 & 37.2 & 35.0  \\
    \hline
    Comp. (ELLA)            & 60.3 & 27.8 & 11.9 &  0.8 & 33.9 & 26.5  \\
    Comp. (Ours)            & 60.8 & 31.7 & 22.4 &  6.7 & 36.8 & 33.9  \\
    \hline
    Student (ELLA)          & 60.8 & 35.6 & 18.6 &  5.0 & 39.1 & 36.2  \\
    Student (Ours)          & 61.1 & \textbf{39.4} & \textbf{27.4} &  \textbf{9.4} & \textbf{41.3} & \textbf{39.3}  \\
    \hline
    LSBN+LUPI~\cite{ExLPose_2023_CVPR}  & 61.1 & 33.7 & 14.7 &  3.4 & 35.3 & 35.1  \\ 
    \hline
    \hline
  \end{tabular}
  \centering
  \caption{
  Full comparison results of ELLA~\cite{ella-eccv} and our method. ``Main'' refers to ``main teacher''. ``Comp.'' refers to ``complementary teacher''. And ``student'' refers to ``student'' distillation model, which is the full model of ELLA.
  The best is \textbf{bold}. 
  }
  \label{tab:comp-ella}
\end{table}
ELLA~\cite{ella-eccv} is based on DEKR~\cite{DEKR2021}, which utilizes different type of loss (e.g. center-offset, joints-tags) for dual-teacher design, while our backbone, ED-Pose~\cite{edpose}, directly regresses to 2D coordinate for each keypoint.
Therefore, we cannot directly use ELLA's dual-teacher in our framework.
In this case, we evaluate our low-light image synthesis in ELLA's dual-teacher pipeline without our proposed DCA. 
In particular, we integrate our synthetic data into the dual-teacher-student distillation framework proposed by ELLA~\cite{ella-eccv}.
\cref{tab:comp-ella} shows the detailed results at each stage of the ELLA framework, including both teacher models and the final student model.

The comparison reveals that while ELLA's main teacher achieves a slightly higher performance on well-lit (WL) images, our main and complementary teachers consistently and significantly outperform their counterparts across all low-light conditions (LL-N, LL-H, and LL-E). 
For instance, our main teacher improves performance on the challenging LL-H and LL-E sets by +9.6 AP and +6.7 AP, respectively.
These results demonstrate that our synthetic data more effectively captures low-light characteristics than ELLA’s handcrafted augmentation, leading to improved teacher models.

Consequently, the stronger teacher models using our synthetic data lead to a more effective student model.
Our final distilled student surpasses the ELLA student by a substantial margin across all low-light subsets on both ExLPose-test and ExLPose-OCN. 
Notably, our student achieves remarkable improvements of +8.8 AP on LL-H, +4.4 AP on LL-E, and +5.1 AP on A7M3. 
These results demonstrate the effectiveness of our approach in generating low-light images for human pose estimation, resulting in a stronger student model within the ELLA framework.

We also include a baseline (LSBN+LUPI~\cite{ExLPose_2023_CVPR}) trained directly with labeled low-light data from ExLPose (dual-camera) as shown in the last row of \cref{tab:comp-ella}. 
Despite using supervised low-light annotations, this baseline is outperformed by our method, indicating that training on synthesized low-light data can generalize better than relying on limited paired low-light data.

\section{Evaluation of the AIN Module}
\begin{table}[!t]
\fontsize{9pt}{10pt}\selectfont 
\setlength{\tabcolsep}{1.2mm} 
\centering
    \begin{tabular}{l| r | rrr | rr}
    \hline
    \hline
          & \multicolumn{6}{c}{\textbf{AP$^{\uparrow}$@{0.5:0.95}}} \\
    \cline{2-7}                 &  WL  & LL-N & LL-H & LL-E & \makecell{A7 \\ M3} & \makecell{RIC \\ OH3} \\
    \hline
    Direct input             & 60.8 & 23.7 &  7.3 &  0.0 & 27.3 & 24.3  \\
    \hline
    Z-score-based norm.     & 60.9 & 25.4 & 13.2 &  2.2 & 28.4 &  25.0  \\
    Fixed factor                & 58.1 & 29.0 & 20.8 &  6.4 & 33.2 &  31.0  \\
    ImageNet-based       & 61.3 & 31.7 & 22.4 &  7.4 & 36.5 &  33.8  \\
    \hline
    Ours  & \textbf{61.5} & \textbf{32.3} & \textbf{23.2} &  \textbf{8.3} & \textbf{37.2} &  \textbf{35.0}  \\
    \hline
    \hline
  \end{tabular}
  \centering
  \caption{
  Evaluation of the AIN module on ExLPose-test and ExLPose-OCN.
  Direct input refers to feeding low-light images into SD without AIN.
  Experiments are conducted using the DEKR pose model~\cite{DEKR2021}, with DHF and LCIM enabled for all normalization approaches.
  The best is \textbf{bold}. 
  }
  \label{tab:abl-ain}
\end{table}

Low-light images often contain extremely low intensity values, which can cause the VAE encoder in the SD model to produce corrupted latent codes.
To address this, we introduce Adaptive Intensity Normalization to the real low-light reference images $I_\text{LL}$ right before feeding it into the SD-VAE encoder.
This process can be formulated as:
\begin{equation}
    I_\text{LL} \leftarrow I_\text{LL} \times \frac{\delta}{\mu_{I_\text{LL}}}
\end{equation}
where $\delta = 0.449$, which is the mean intensity of ImageNet~\cite{deng2009imagenet} across all channels, and $\mu_{I_\text{LL}}$ represents average intensity of $I_\text{LL}$ across all channels.
We conduct a comprehensive ablation study to validate the AIN module with DEKR~\cite{DEKR2021} as the pose estimation model. As shown in~\cref{tab:abl-ain}, we compare our method against several alternative normalization strategies.

First, we establish a baseline by feeding low-light images directly into the network without any normalization (``Direct input''). This approach yields poor performance, with Average Precision (AP) scores dropping to a mere 7.3 on the LL-H set and 0.0 on the LL-E set. This result underscores the critical need for an effective input normalization technique to handle the challenges of low-light conditions.

\begin{table}[!t]
\fontsize{9pt}{10pt}\selectfont 
\setlength{\tabcolsep}{0.9mm} 
\centering
    \begin{tabular}{l| r | rrr | rr | r }
    \hline
    \hline
          & \multicolumn{7}{c}{\textbf{AP$^{\uparrow}$@{0.5:0.95}}} \\
    \cline{2-8}         &  WL  & LL-N & LL-H & LL-E & \makecell{A7 \\ M3} & \makecell{RIC \\ OH3} & \makecell{EHPT \\ -XC} \\
    \hline
    $z_0$  & 66.8 & 29.7 & 12.1 & 0.1 & 40.0 & 36.8 & 11.3 \\
    $z_0,z_1$              & 66.8 & 31.4 & 19.2 &  2.4 & 41.4 &  37.4 & 15.4 \\
    $z_0,z_1,z_2$          & 67.2 & 35.3 & 23.2 &  5.8 & 43.8 &  39.9 & 26.2 \\
    $z_0,z_1,z_2,z_3$      & \textbf{67.4} & 37.7 & 26.5 &  7.7 & 47.9 &  43.7 & 29.7 \\
    $z_0,z_1,z_2,z_3,z_4$  & 67.3 & \textbf{38.7} & \textbf{28.0} &  \textbf{11.7} & \textbf{55.0} & \textbf{47.9} & \textbf{31.0} \\
    \hline
    \hline
  \end{tabular}
  \centering
  \caption{
  Ablation study of LCIM on ExLPose-test, ExLPose-OCN, and EHPT-XC.
  $z_0$ refers to baseline SD without any extra intermediate features.
  $z_1$ to $z_4$ represent low-to-high-frequency information fused in a coarse-to-fine integration strategy.
  Results are reported with AIN, DHF and DCA.
  The best is \textbf{bold}.
  }
  \label{tab:abl-skip}
\end{table}

Next, we evaluate several alternative normalization strategies. 
Applying z-score-based normalization offers only a marginal improvement, which is formulated as
\begin{equation}
    I'_\text{LL} = \frac{\sigma_\text{ImageNet}}{\sigma_\text{LL}} (I_\text{LL} - \mu_\text{LL}) + \mu_\text{ImageNet}
\end{equation}
\noindent This approach is not suitable for low-light images, where pixel values are highly concentrated near zero. The mean-subtraction operation introduces numerous negative values, which can disrupt the original signal distribution and discard subtle but important low-light noise characteristics. Using a fixed scaling factor for the whole dataset is another option but not optimal as well, which is formulated as
\begin{equation}
    I'_\text{LL} = I_\text{LL} \times k
\end{equation}
\noindent Since low-light scenes exhibit diverse illumination levels, a single fixed factor can cause over-exposure in relatively brighter images and insufficient enhancement in darker ones, failing to produce a consistently normalized input. 
A third option, per-channel scaling (e.g., using ImageNet's standard ``[0.485, 0.456, 0.406]'' values), provides a slightly better result. 
However, this approach distorts the intrinsic color balance by altering the relative strengths of the R, G, and B channels. This can cause an undesirable color shift and prevent the model from learning to handle realistic low-light color noise faithfully.

In contrast, our proposed AIN, which adaptively rescales each image using a single, content-aware factor, achieves superior performance across all evaluated scenarios as shown in Table~\ref{tab:abl-ain}.
AIN improves the AP to 32.3, 23.2, and 8.3 on LL-N, LL-H, and LL-E, respectively, outperforming all other variants. 
By preserving the inter-channel ratios, our method avoids color distortion. By adapting the scaling factor to each image's mean intensity, it effectively normalizes brightness without introducing clipping artifacts. 
This process provides a stable and informative input for the downstream network, leading to significantly improved human pose estimation accuracy. 
These results validate our design choices and demonstrate the effectiveness of AIN for low-light human pose estimation.



\section{Analysis of LCIM}
\label{sec:ablation_skips}
We now analyze the core of our LCIM module: the multi-scale intermediate features. As detailed in~\cref{tab:abl-skip}, we start with a baseline model ($z_0$) that omits all intermediate features, then progressively integrate features from coarse to fine levels ($z_1$ to $z_4$). 
The $z_0$ model, achieves a modest 40.0 AP on A7M3 and 36.8 AP on RICOH3.

These results show that fusing multi-scale features is important. 
Each added intermediate feature brings a consistent performance improvement. 
Adding all four feature levels ($+z_1+z_2+z_3+z_4$) results in our strongest model, improving the AP by +15.0 on A7M3 (40.0$\rightarrow$55.0) and +11.1 on RICOH3 (36.8$\rightarrow$47.9) compared to the $z_0$ baseline. 
This analysis shows that our coarse-to-fine fusion strategy effectively uses multi-scale latent features from the SD encoder, which is important for robust pose estimation under challenging low-light conditions.

\begin{figure*}[t]
  \centering
  \begin{subfigure}[t]{0.3\linewidth}
    \centering
    \includegraphics[width=\linewidth]{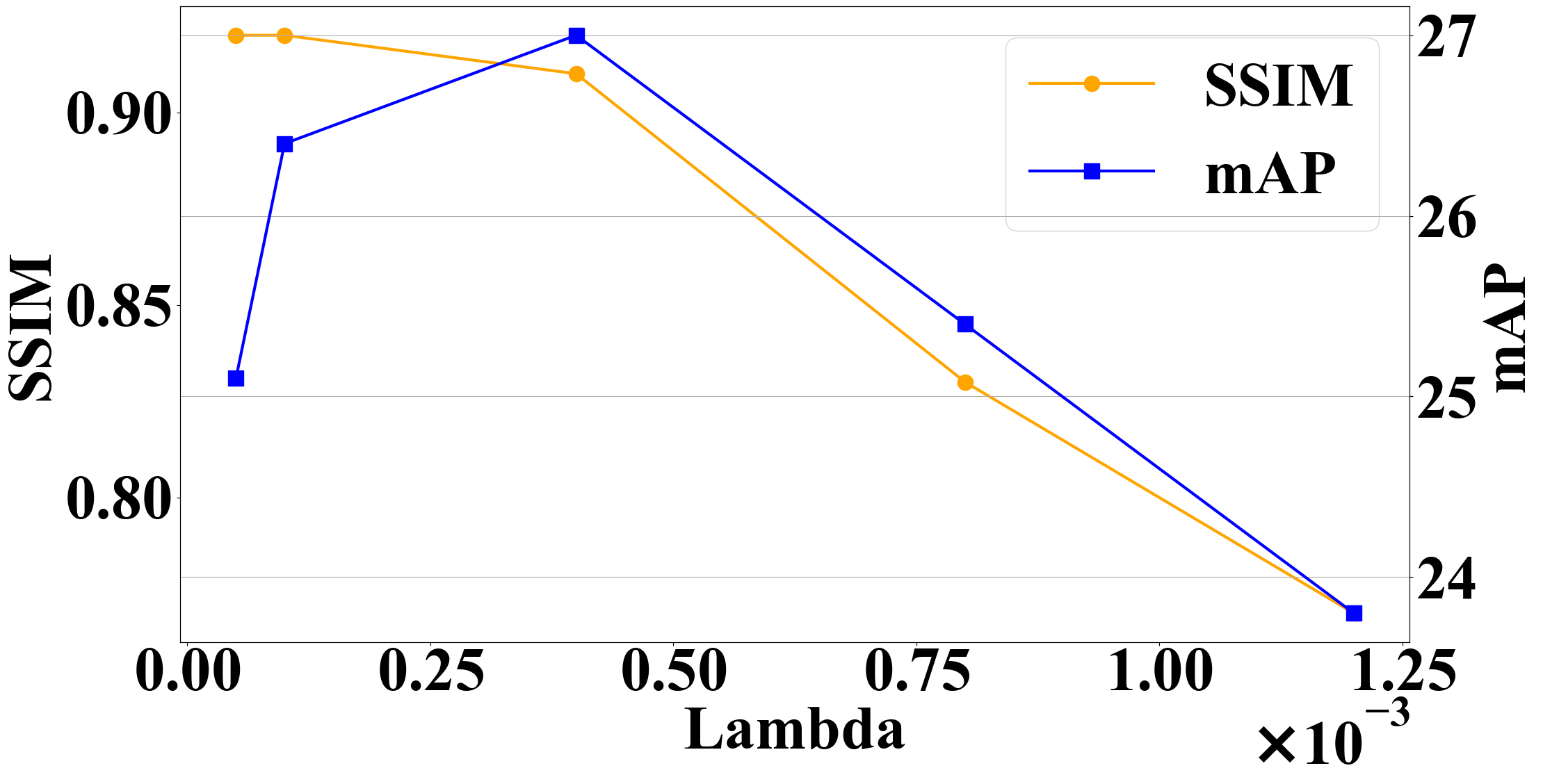}
    \caption{
    }
    \label{fig:lambda}
  \end{subfigure}
  \hspace{3cm}
  \begin{subfigure}[t]{0.3\linewidth}
    \centering
    \includegraphics[width=\linewidth]{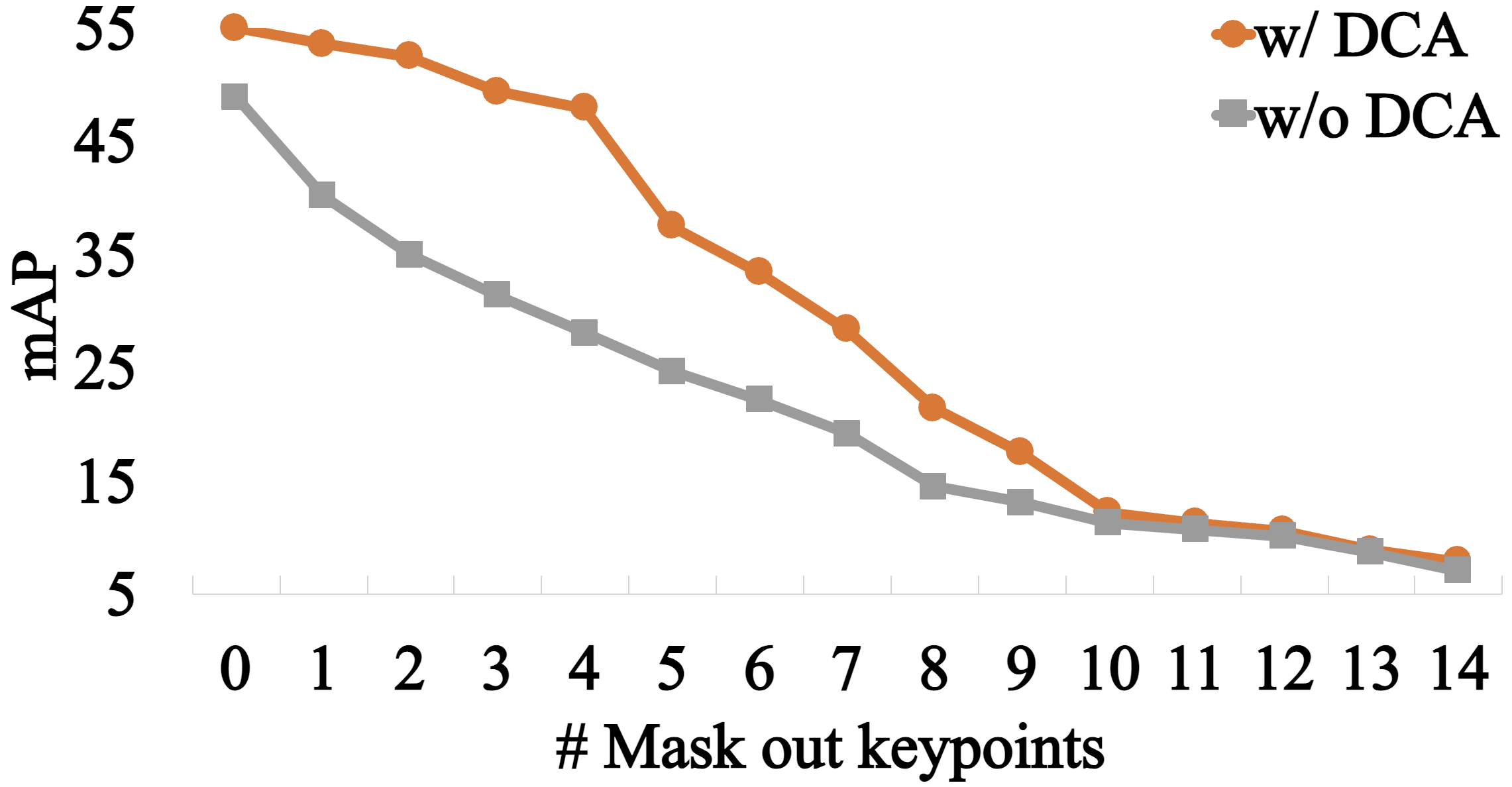}
    \caption{
    }
    \label{fig:mask}
  \end{subfigure}
  \vspace{-0.1cm}
  \caption{(a) Effect of $\lambda$. (b) Masking evaluation w/ and w/o DCA.}
  \label{fig:mask-out}
  \vspace{-0.3cm}
\end{figure*}

\section{Evaluation of $\lambda$ and DCA}
\subsection{Sensitivity of $\lambda$}
As defined in~\cref{equ:llcd_loss_revised} of the main paper, $\lambda$ controls the relative weight between the pixel-level MSE loss $\mathcal{L}_{\text{MSE}}$ and the frequency-domain loss $\mathcal{L}_{\text{freq}}$ during LCIM training.
We analyze its effect by varying $\lambda$ and measuring both image-level quality (SSIM between synthesized and real low-light images) and downstream pose estimation performance (mAP on ExLPose-OCN).
As shown in~\cref{fig:lambda}, increasing $\lambda$ places more emphasis on high-frequency detail preservation, which improves the fidelity of low-light noise patterns in the synthesized images and leads to higher mAP.
However, beyond a certain point, an overly large $\lambda$ degrades content consistency, as indicated by a drop in SSIM.
This occurs because the frequency loss begins to dominate, causing the decoder to prioritize noise texture over structural content from the well-lit source image.
Conversely, a small $\lambda$ underweights the frequency loss, producing synthesized images that lack realistic low-light noise and thus provide insufficient training signal for the pose estimator.
Based on this tradeoff, we set $\lambda = 4 \times 10^{-4}$ in all experiments.

\subsection{Robustness Analysis of DCA}
To further evaluate DCA beyond the ablation study in the main paper, we design a masking experiment that probes DCA's ability to use pose priors when visual information is missing.
Specifically, we evaluate on ExLPose-OCN (A7M3) by progressively masking a random subset of ground-truth keypoints in each test image, simulating scenarios where varying numbers of joints are occluded or invisible.
As shown in~\cref{fig:mask}, DCA consistently outperforms the baseline (without DCA) when a small number of keypoints are masked.
This is because DCA detects unreliable image cues for the masked keypoints and shifts its reliance toward learned pose priors, leading to more reliable predictions for these keypoints compared to relying on noisy visual cues alone.
As the number of masked keypoints increases, the gap between DCA and the baseline narrows.
This is expected: when the majority of keypoints are invisible, even pose priors offer limited information, as the model has fewer visible joints to anchor its structural reasoning.
This also indicates a limitation of DCA under extreme conditions, where very limited visual evidence constrains the effectiveness of pose priors.

\begin{table}[t]
\fontsize{9pt}{10pt}\selectfont 
\setlength{\tabcolsep}{1mm} 
\centering
    \begin{tabular}{l| c | rrr | cc | c}
    \hline
    \hline
          & \multicolumn{7}{c}{\textbf{AP$^{\uparrow}$@{0.5:0.95}}} \\
    \cline{2-8}            &  WL  & LL-N & LL-H & LL-E & \makecell{A7 \\ M3} & \makecell{RIC \\ OH3} & \makecell{EHPT \\ -XC} \\
    \hline
    SE-Block~\cite{se-block}     & 62.4 & 36.7 & 26.3 & 9.5 & 50.3 & 46.5 & 26.7  \\
    CBAM~\cite{cbam}         & 62.5 & 37.0 & 26.2 & 9.8 & 51.1 & 46.2 & 27.0  \\
    \hline
    Ours (DCA)        & \textbf{67.3} & \textbf{38.7} & \textbf{28.0} & \textbf{11.7} & \textbf{55.0} & \textbf{47.9} & \textbf{31.0}  \\

    \hline
    \hline
  \end{tabular}
  \centering
  \caption{
    Comparison of SE-Block~\cite{se-block}, CBAM~\cite{cbam}, and our DCA gating mechanism. The best is \textbf{bold}.
  }
  \label{tab:abl-dca}
\end{table}

\begin{table}[t]
\fontsize{9pt}{10pt}\selectfont 
\setlength{\tabcolsep}{1mm} 
\centering
    \begin{tabular}{l| c | rrr | cc | c}
    \hline
    \hline
          & \multicolumn{7}{c}{\textbf{AP$^{\uparrow}$@{0.5:0.95}}} \\
    \cline{2-8}            &  WL  & LL-N & LL-H & LL-E & \makecell{A7 \\ M3} & \makecell{RIC \\ OH3} & \makecell{EHPT \\ -XC} \\
    \hline
    ControlNet~\cite{controlnet}   & 66.3 & 31.7 & 16.4 & 2.7 & 47.6 & 43.7 & 22.4  \\
    IP-Adapter~\cite{ip-adapter}  & 65.8 & 31.5 & 17.1 & 3.5 & 48.4 & 43.4 & 24.1  \\
    \hline
    Ours         & \textbf{67.3} & \textbf{38.7} & \textbf{28.0} & \textbf{11.7} & \textbf{55.0} & \textbf{47.9} & \textbf{31.0}  \\

    \hline
    \hline
  \end{tabular}
  \centering
  \caption{
    Comparison of ControlNet~\cite{controlnet}, IP-Adapter~\cite{ip-adapter}, and our method. The best is \textbf{bold}.
  }
  \label{tab:controlnet}
\end{table}

\section{Ablation Study of DCA}
We compare DCA against two general-purpose attention gating mechanisms: SE-Block~\cite{se-block} and CBAM~\cite{cbam}.
Each replaces DCA at the same position in the decoder layer, fusing $\mathbf{Q}_\text{pose}$ and $\mathbf{Q}_\text{image}$ before the FFN.
All three variants use the same synthesized low-light training data, with DHF and LCIM enabled for all variants.
As shown in~\cref{tab:abl-dca}, SE-Block and CBAM both improve over the no-gating baseline (Table~5 in the main paper, ``+ DHF'' row).
However, DCA outperforms both.
On ExLPose-test, DCA leads SE-Block by 1.7--2.2 AP across the low-light subsets and by 4.9 AP on well-lit images.
The gap is larger on cross-dataset evaluation, where DCA exceeds SE-Block and CBAM by 4.3 and 4.0 AP on EHPT-XC, respectively.
This is likely due to the explicit softmax competition in DCA between pose-prior and image-cue channels: the two weights sum to one per keypoint, forcing the model to make a binary-like choice for each joint.
SE-Block and CBAM instead learn generic channel or spatial reweighting without this structural constraint, so they lack the inductive bias to suppress unreliable image cues for specific keypoints.

\begin{figure*}[ht]

\newlength{\tablelength}
\setlength{\tablelength}{0.145\linewidth}

\newlength{\imgwidth}
\setlength{\imgwidth}{0.16\linewidth}

\centering

    \begin{minipage}[c]{\tablelength}
        \resizebox{\linewidth}{!}{
            %
            \centering
            \renewcommand{\arraystretch}{1.6}
            \begin{tabular}{r|c|c}
                 & \textbf{Before} & \textbf{After} \\
                \hline
                \hline
                L. shoulder & \val{1.73} & \val{1.74} \\
                R. shoulder & \val{1.81} & \val{1.78} \\
                L. elbow & \val{1.72} & \val{1.41} \\
                R. elbow & \val{1.77} & \val{1.47} \\
                L. wrist & \val{1.69} & \val{1.24} \\
                R. wrist & \val{1.74} & \val{1.49} \\
                L. hip & \val{1.70} & \val{1.72} \\
                R. hip & \val{1.80} & \val{1.81} \\
                L. knee & \val{1.85} & \val{1.55} \\
                R. knee & \val{1.84} & \val{1.57} \\
                L. ankle & \val{1.76} & \val{1.51} \\
                R. ankle & \val{1.81} & \val{1.52} \\
                Head & \val{1.89} & \val{1.93} \\
                Neck & \val{1.82} & \val{1.89} \\
                \hline
                \hline
            \end{tabular}
        }
    \end{minipage}
    \begin{minipage}[c]{\imgwidth}
        \caption*{Before}
        \includegraphics[
            trim={360 0 360 0}, 
            clip, 
            width=\textwidth
        ]{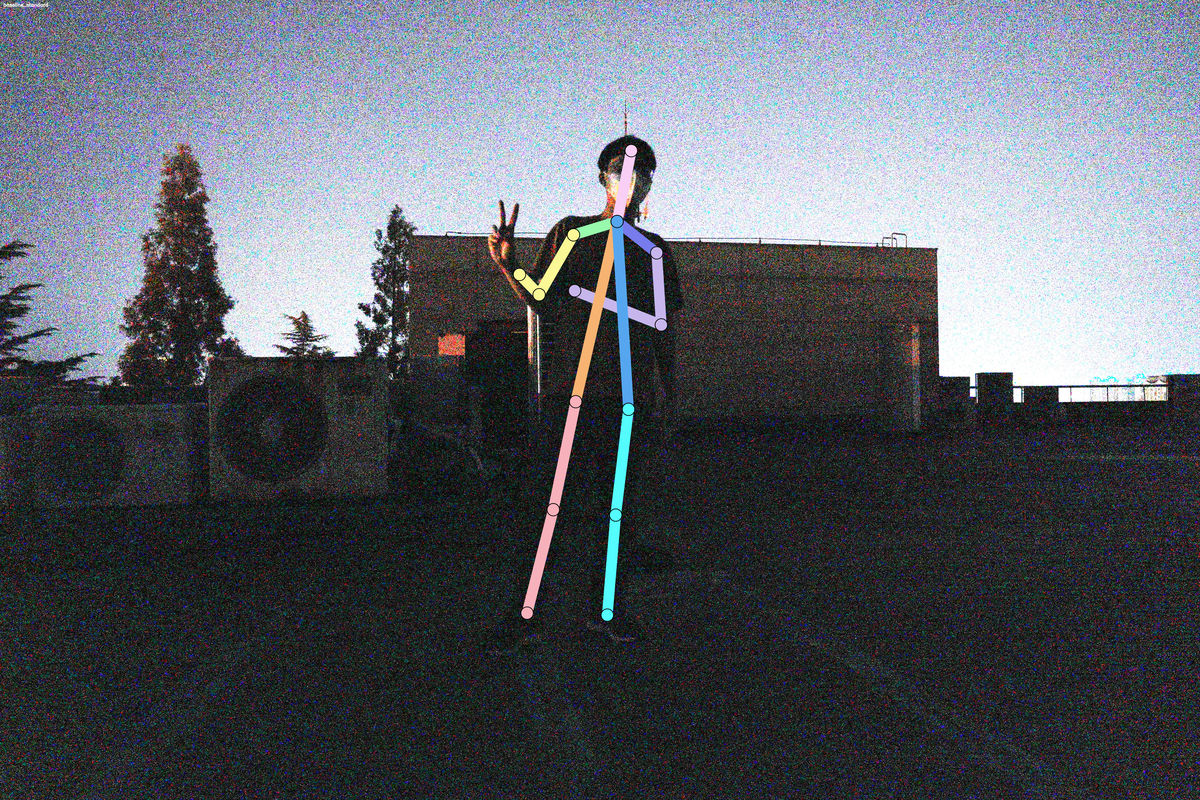}
    \end{minipage}
    \begin{minipage}[c]{\imgwidth}
        \caption*{After}
        \includegraphics[
            trim={360 0 360 0}, 
            clip, 
            width=\textwidth
        ]{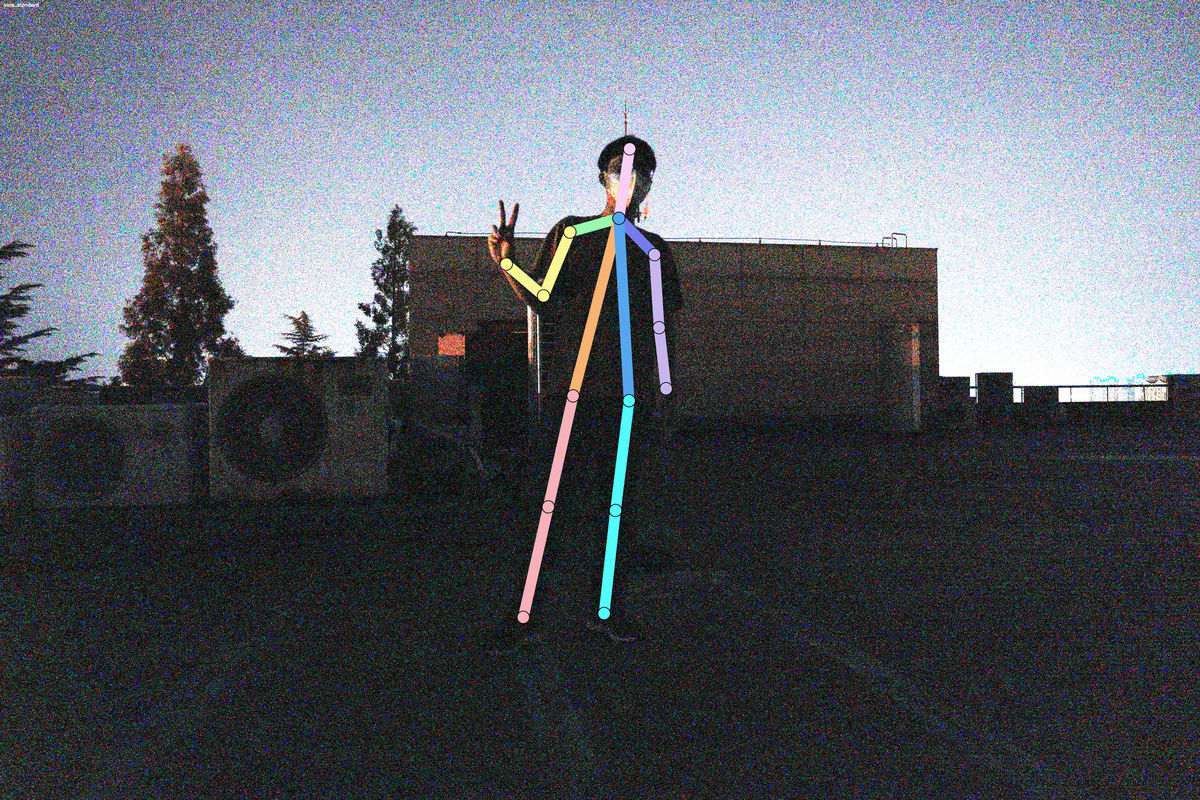}
    \end{minipage}
    \hfill
    \begin{minipage}[c]{\tablelength}
        \resizebox{\linewidth}{!}{
            %
            \centering
            \renewcommand{\arraystretch}{1.6}
            \begin{tabular}{r|c|c}
                 & \textbf{Before} & \textbf{After} \\
                \hline
                \hline
                L. shoulder & \val{1.74} & \val{1.71} \\
                R. shoulder & \val{1.71} & \val{1.74} \\
                L. elbow & \val{1.82} & \val{1.73} \\
                R. elbow & \val{1.81} & \val{1.72} \\
                L. wrist & \val{1.85} & \val{1.74} \\
                R. wrist & \val{1.84} & \val{1.93} \\
                L. hip & \val{1.91} & \val{1.75} \\
                R. hip & \val{1.93} & \val{1.73} \\
                L. knee & \val{1.83} & \val{1.13} \\
                R. knee & \val{1.86} & \val{1.17} \\
                L. ankle & \val{1.88} & \val{1.31} \\
                R. ankle & \val{1.87} & \val{1.35} \\
                Head & \val{1.79} & \val{1.84} \\
                Neck & \val{1.82} & \val{1.70} \\
                \hline
                \hline
            \end{tabular}
        }
    \end{minipage}
    \begin{minipage}[c]{\imgwidth}
        \caption*{Before}
        \includegraphics[
            trim={0 0 720 0},
            clip,
            width=\textwidth
        ]{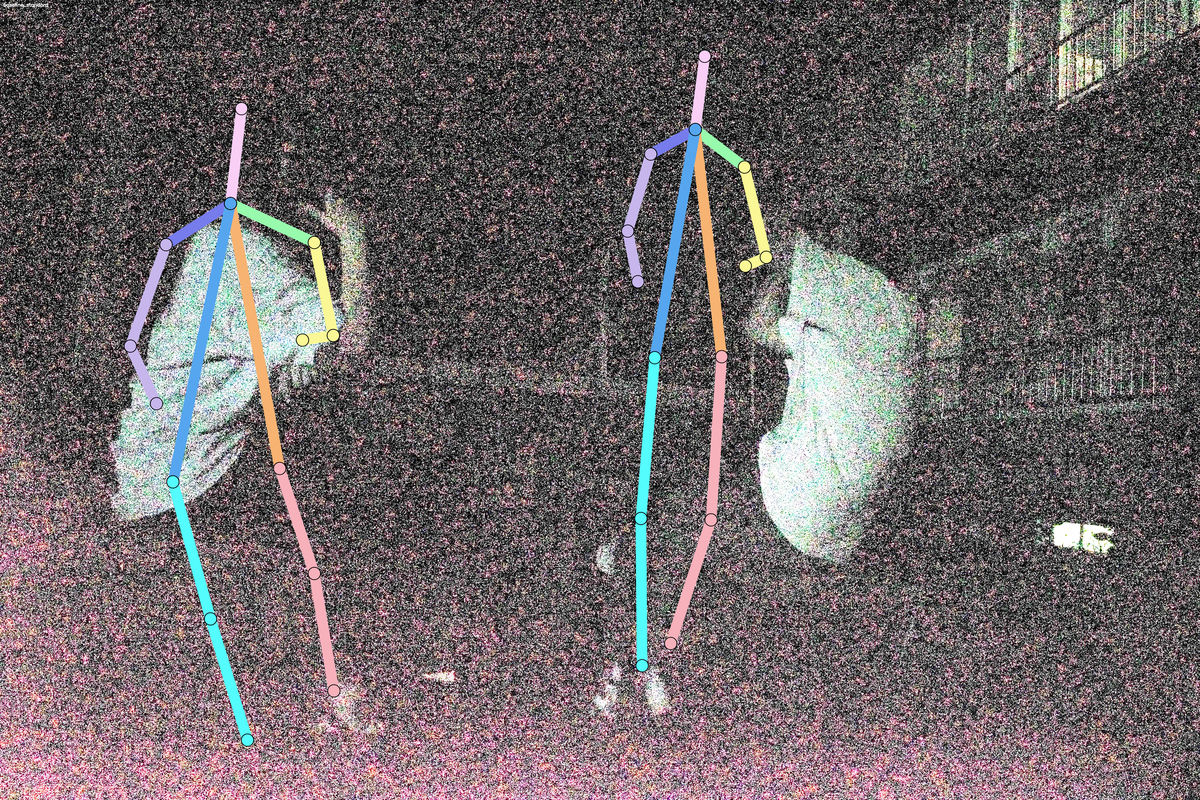}
    \end{minipage}
    \begin{minipage}[c]{\imgwidth}
        \caption*{After}
        \includegraphics[
            trim={0 0 720 0},
            clip,
            width=\textwidth
        ]{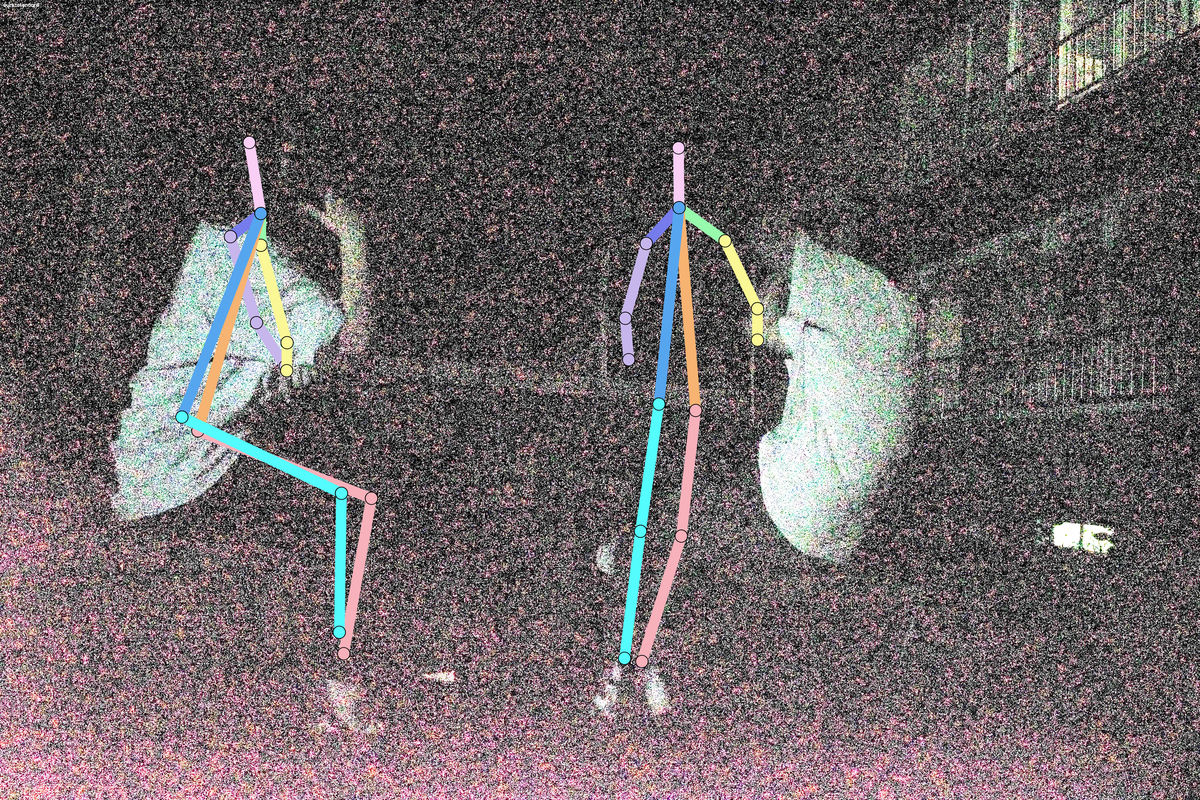}
    \end{minipage}
    \caption{Qualitative ablation of our DCA module. L. represents left, R. represents right.}
    \label{fig:dca-qual}
\end{figure*}

\section{Comparison to ControlNet and IP-Adapter}

To further evaluate the quality of our synthesized low-light data, we compare against two commonly used diffusion-based conditioning methods: ControlNet~\cite{controlnet} and IP-Adapter~\cite{ip-adapter}.
Both methods are trained on paired well-lit and low-light images from the ExLPose dual-camera system, providing them with direct pixel-level supervision that our method does not require.
ControlNet adds spatial conditioning to the diffusion model, while IP-Adapter injects reference image features through a decoupled cross-attention mechanism.
All experiments are conducted using ED-Pose~\cite{edpose} with the DCA module enabled, and our method also includes the proposed DHF and LCIM.

As shown in~\cref{tab:controlnet}, our method outperforms both baselines across all evaluation sets without relying on paired training data.
On ExLPose-test, the performance gap widens as conditions become more challenging: our method leads by 7.0 AP on LL-N, 10.9 AP on LL-H, and 8.2 AP on LL-E compared to the best-performing baseline.
On ExLPose-OCN, we observe gains of 6.6 AP on A7M3 and 4.2 AP on RICOH3.
On the cross-dataset EHPT-XC benchmark, our method achieves 31.0 AP, surpassing IP-Adapter by 6.9 AP.
This advantage mainly comes from our DHF and LCIM modules, which extract and inject high-frequency low-light characteristics at multiple scales in the decoder.
In contrast, ControlNet and IP-Adapter use general-purpose conditioning mechanisms that are not designed for modeling low-light noise patterns.
These results suggest that task-specific characteristic injection can be more effective than general diffusion-based conditioning for low-light data synthesis, even when the latter has access to paired supervision.

\section{Qualitative results}

\subsection{Comparison of Pose Prediction}
We present a qualitative comparison of pose prediction results between our proposed UDAPose and related methods including DarkIR~\cite{feijoo2025darkir}, QuadPrior~\cite{quadprior}, CycleGAN~\cite{CycleGAN2017}, UNSB~\cite{kim2023unsb}, EnCo~\cite{cai2024enco}, ELLA~\cite{ella-eccv} in~\cref{fig:pose-qual}.
The qualitative results clearly demonstrate the superior capability of our approach in predicting human pose under low-light conditions.
We first observe the limitations of enhancement-based methods. Both DarkIR~\cite{feijoo2025darkir} and QuadPrior~\cite{quadprior} rely on a pre-processing step to enhance the image.
However, this enhancement procedure is often ill-posed in extreme darkness and can introduce visual artifacts that mislead the subsequent pose estimation model.
This is evident as they produce a biologically implausible pose, or fail to detect the person altogether.
In contrast, domain adaptation methods such as CycleGAN~\cite{CycleGAN2017}, UNSB~\cite{kim2023unsb}, EnCo~\cite{cai2024enco}, and ELLA~\cite{ella-eccv} show improved performance by training on synthetic data.
Nevertheless, they still produce inaccurate joint locations.
We attribute this to the limited fidelity of their synthetic data, which fails to fully capture the complex degradations of real-world low-light imagery.
Our method overcomes these limitations and yields a substantially more accurate result.
This superior performance stems from two key factors: (1) our high-fidelity data synthesis pipeline, which provides training examples that reflect low-light characteristics, and (2) our DCA module, which adaptively balances unreliable visual cues from the noisy image with robust, learned anatomical priors.
This allows our model to maintain structural coherence and precision even in extreme conditions.

\subsection{Comparison of Synthesized Images}
We present a qualitative comparison between our proposed UDAPose and related methods including QuadPrior~\cite{quadprior}, CycleGAN~\cite{CycleGAN2017}, UNIT~\cite{UNIT}, UNSB~\cite{kim2023unsb}, StyleID~\cite{chung2024style} in~\cref{fig:syn-qual}. 
The qualitative results clearly demonstrate the superior capability of our approach in generating low-light images that capture the characteristics of low-light images.

Among the image enhancement-based methods, QuadPrior~\cite{quadprior} attempts to brighten low-light images but tends to produce over-smoothed results with significant loss of texture and detail.
For the image-to-image translation methods, CycleGAN~\cite{CycleGAN2017} produces images with excessive color shifts and unrealistic noise patterns. 
UNIT~\cite{UNIT} and UNSB~\cite{kim2023unsb} generate images with irregular noise distributions that significantly differ from real low-light conditions, making them less effective for training human pose estimation models. 
StyleID~\cite{chung2024style}, while better at preserving the overall scene structure, still struggles to accurately capture the complex noise patterns of the low-light images.

\newcommand{\examplePose}[1]{
    \begin{figure*}[!ht]
        \centering

        \pgfmathtruncatemacro{\remainder}{mod(\value{posefigurecounter}, 11)}

        \ifnum\value{posefigurecounter}=1
            \caption{Pose predictions of UDAPose compared against competing methods. The first two columns show results from enhancement-based methods; all other columns display results on the original low-light images. The low-light images are scaled for visualization only.}
            \label{fig:pose-qual}
        \else
        \fi

        \begin{minipage}[c]{0.138\linewidth}
            \includegraphics[width=\linewidth]{figures/pose_qual/ll_downsample_s30_q80/darkir/#1.jpg}
        \end{minipage}
        \begin{minipage}[c]{0.138\linewidth}
            \includegraphics[width=\linewidth]{figures/pose_qual/ll_downsample_s30_q80/quadprior/#1.jpg}
        \end{minipage}
        \begin{minipage}[c]{0.138\linewidth}
            \includegraphics[width=\linewidth]{figures/pose_qual/ll_downsample_s30_q80/cyclegan_standard/#1.jpg}
        \end{minipage}
        \begin{minipage}[c]{0.138\linewidth}
            \includegraphics[width=\linewidth]{figures/pose_qual/ll_downsample_s30_q80/unsb_standard/#1.jpg}
        \end{minipage}
        \begin{minipage}[c]{0.138\linewidth}
            \includegraphics[width=\linewidth]{figures/pose_qual/ll_downsample_s30_q80/enco_standard/#1.jpg}
        \end{minipage}
        \begin{minipage}[c]{0.138\linewidth}
            \includegraphics[width=\linewidth]{figures/pose_qual/ll_downsample_s30_q80/ella_standard/#1.jpg}
        \end{minipage}
        \begin{minipage}[c]{0.138\linewidth}
            \includegraphics[width=\linewidth]{figures/pose_qual/ll_downsample_s30_q80/ours_standard/#1.jpg}
        \end{minipage}

        \begin{minipage}[c]{0.138\linewidth}
		    \centerline{\scriptsize{DarkIR~\cite{feijoo2025darkir}}}
        \end{minipage}
        \begin{minipage}[c]{0.138\linewidth}
		    \centerline{\scriptsize{QuadPrior~\cite{quadprior}}}
        \end{minipage}
        \begin{minipage}[c]{0.138\linewidth}
		    \centerline{\scriptsize{CycleGAN~\cite{CycleGAN2017}}}
        \end{minipage}
        \begin{minipage}[c]{0.138\linewidth}
		    \centerline{\scriptsize{UNSB~\cite{kim2023unsb}}}
        \end{minipage}
                \begin{minipage}[c]{0.138\linewidth}
		    \centerline{\scriptsize{EnCo~\cite{cai2024enco}}}
        \end{minipage}
                \begin{minipage}[c]{0.138\linewidth}
		    \centerline{\scriptsize{ELLA~\cite{ella-eccv}}}
        \end{minipage}
                \begin{minipage}[c]{0.138\linewidth}
		    \centerline{\scriptsize{Ours}}
        \end{minipage}
    \end{figure*}
}

\newcommand{\poseimageids}{913, 1299, 1484, 523, 1395, 1294, 1486, 550, 1305, 1318, 1579}

\newlength{\originaltextheight}
\newlength{\originaltopmargin}
 
\newcommand{\adjustVerticalMargins}[2]{%
    \setlength{\originaltextheight}{\textheight}%
    \setlength{\originaltopmargin}{\topmargin}%
 
    \clearpage
 
    \setlength{\topmargin}{#1 - 1in - \headheight - \headsep}%
 
    \setlength{\textheight}{\paperheight - #1 - #2}%
 
 
}
 
\newcommand{\restoreVerticalMargins}{%
    \clearpage
    \setlength{\textheight}{\originaltextheight}%
    \setlength{\topmargin}{\originaltopmargin}%
 
}
\newcounter{posefigurecounter}

\adjustVerticalMargins{1.0in}{0.5in}

\setcounter{posefigurecounter}{0}
\foreach \id in \poseimageids{
    \stepcounter{posefigurecounter} 
    \examplePose{\id}
}

\restoreVerticalMargins 

\newcommand{\exampleSyn}[1]{
    \begin{figure*}[!ht]
        \centering

        \pgfmathtruncatemacro{\remainder}{mod(\value{figurecounter}, 4)}

        \ifnum\value{figurecounter}=1
            \caption{Qualitative comparison of our data synthesis method with baselines.}
            \label{fig:syn-qual}
        \else
        \fi

        \begin{minipage}[c]{0.238\linewidth}
            \includegraphics[width=\linewidth]{figures/collect_downsample_s32/#1/wl.png}
        \end{minipage}
        \begin{minipage}[c]{0.238\linewidth}
            \includegraphics[width=\linewidth]{figures/collect_downsample_s32/#1/gt_ll_04.png}
        \end{minipage}
        \begin{minipage}[c]{0.238\linewidth}
            \includegraphics[width=\linewidth]{figures/collect_downsample_s32/#1/quadprior.png}
        \end{minipage}
        \begin{minipage}[c]{0.238\linewidth}
            \includegraphics[width=\linewidth]{figures/collect_downsample_s32/#1/cyclegan.png}
        \end{minipage}

        \begin{minipage}[c]{0.238\linewidth}
		    \centerline{\scriptsize{Well-lit}}
        \end{minipage}
        \begin{minipage}[c]{0.238\linewidth}
		    \centerline{\scriptsize{Paired Low-light}}
        \end{minipage}
        \begin{minipage}[c]{0.238\linewidth}
		    \centerline{\scriptsize{QuadPrior~\cite{quadprior}}}
        \end{minipage}
        \begin{minipage}[c]{0.238\linewidth}
		    \centerline{\scriptsize{CycleGAN~\cite{CycleGAN2017}}}
        \end{minipage}

        \begin{minipage}[c]{0.238\linewidth}
            \includegraphics[width=\linewidth]{figures/collect_downsample_s32/#1/unit.png}
        \end{minipage}
        \begin{minipage}[c]{0.238\linewidth}
            \includegraphics[width=\linewidth]{figures/collect_downsample_s32/#1/unsb.png}
        \end{minipage}
        \begin{minipage}[c]{0.238\linewidth}
            \includegraphics[width=\linewidth]{figures/collect_downsample_s32/#1/styleid.png}
        \end{minipage}
        \begin{minipage}[c]{0.238\linewidth}
            \includegraphics[width=\linewidth]{figures/collect_downsample_s32/#1/our_v04.png}
        \end{minipage}

        \begin{minipage}[c]{0.238\linewidth}
		    \centerline{\scriptsize{UNIT~\cite{UNIT}}}
        \end{minipage}
        \begin{minipage}[c]{0.238\linewidth}
		    \centerline{\scriptsize{UNSB~\cite{kim2023unsb}}}
        \end{minipage}
        \begin{minipage}[c]{0.238\linewidth}
		    \centerline{\scriptsize{StyleID~\cite{chung2024style}}}
        \end{minipage}
        \begin{minipage}[c]{0.238\linewidth}
		    \centerline{\scriptsize{Ours}}
        \end{minipage}

    \end{figure*}
}

\newcommand{\imageids}{2262, 1817, 2310, 2117, 629, 279, 563, 512, 453, 460, 495, 507, 390, 433, 1835, 1617}

\newcounter{figurecounter}

\adjustVerticalMargins{0.8in}{0.45in}

\setcounter{figurecounter}{0}
\foreach \id in \imageids{
    \stepcounter{figurecounter} 
    \exampleSyn{\id}
}

\restoreVerticalMargins

\clearpage
\adjustVerticalMargins{1.0in}{1.125in}
\restoreVerticalMargins

In contrast, our method generates low-light images that exhibit low-light noise characteristics.
The synthetic images produced by our method preserve high-frequency details while modeling complex noise patterns observed in low-light conditions. 
The LCIM module is key to the superior quality of our synthetic low-light images, as it effectively captures and transfers complex low-light characteristics from unpaired real low-light images to well-lit inputs. 
As a result, our UDAPose overcomes the limitations of existing approaches, generating more effective training data that better prepares the pose estimation model for low-light scenarios.

\subsection{Comparison of DCA}
We provide a qualitative comparison to demonstrate the effectiveness of our DCA module.
Without DCA, the model tends to assign uniformly high importance to image cues for all keypoints, as indicated by the consistently high values in the ``Before'' columns.
This forces the model to overly rely on visual evidence, even when it is corrupted by noise or low visibility.
Consequently, this leads to erroneous human pose predictions as shown in~\cref{fig:dca-qual}. 
Our DCA module effectively resolves this issue by learning to dynamically balance the influence of image cues and pose priors.
As shown in the ``After'' columns, DCA significantly reduces the cue weights for keypoints with low visibility.
By down-weighting these unreliable signals, the model can leverage its learned pose priors for improved pose estimation.
This results in substantially more accurate and coherent poses, correcting the initial errors and demonstrating that DCA is crucial for achieving robustness in challenging, low-visibility conditions.

\section{Limitations}
\label{sec:limitations}
While our results are promising, there are still opportunities to build on this work in future research.
The current framework, including the proposed LCIM, DHF, and DCA modules, is specifically tailored to model degradations from insufficient illumination, primarily by transferring noise characteristics and balancing unreliable visual cues with learned pose priors. 
A promising future direction is to extend this generative approach to handle a broader spectrum of low-visibility scenarios, such as dense fog, heavy rain, or severe motion blur. This would likely require designing new modules capable of synthesizing these more complex degradations, thereby enhancing the model's generalization to diverse and challenging real-world conditions.

The reliance on a large-scale diffusion model like SD introduces substantial computational overhead. The data synthesis pipeline is resource-intensive, requiring significant GPU memory and time for generating the training dataset. This presents a practical barrier to rapid adaptation for new, custom low-light environments. Future work could explore the use of more efficient generative models, such as consistency models or distilled diffusion models, to mitigate this cost and improve accessibility.

%

%

\end{document}